\documentclass{article}

\usepackage[final]{neurips_2025}

\usepackage[utf8]{inputenc} 
\usepackage[T1]{fontenc}    
\usepackage{hyperref}       
\usepackage{url}            
\usepackage{booktabs}       
\usepackage{caption}
\usepackage{tabularx}
\usepackage{multirow}
\usepackage{enumitem}
\usepackage{amsfonts}  
\usepackage{makecell}
\usepackage{graphicx}  
\usepackage{nicefrac}       
\usepackage{microtype}      
\usepackage{xcolor}         
\usepackage{amsmath}
\usepackage{amssymb}
\usepackage{algorithm}
\usepackage{algorithmic}
\newcommand{\methodname}{{\tt{FedCBDR}}}

\title{Class-wise Balancing Data Replay for Federated Class-Incremental Learning}

\author{
  \textbf{Zhuang Qi}$^{1}$,
  \textbf{Ying-Peng Tang}$^{2}$,
  \textbf{Lei Meng}$^{1,}$\thanks{Corresponding author}\ ,
  \textbf{Han Yu}$^{2}$,
  \textbf{Xiaoxiao Li}$^{3,4}$,
  \textbf{Xiangxu Meng}$^{1}$\\
  $^{1}$School of Software, Shandong University, China \\
  $^{2}$College of Computing and Data Science, Nanyang Technological University, Singapore \\
  $^{3}$Department of Electrical and Computer Engineering, University of British Columbia, Canada \\
  $^4$ Vector Institute, Canada\\
  \texttt{z\_qi@mail.sdu.edu.cn, yingpeng.tang@ntu.edu.sg, lmeng@sdu.edu.cn,} \\
  \texttt{han.yu@ntu.edu.sg, xiaoxiao.li@ece.ubc.ca, mxx@sdu.edu.cn}
}

\begin{document}

\maketitle

\begin{abstract}
Federated Class Incremental Learning (FCIL) aims to collaboratively process continuously increasing incoming tasks across multiple clients. Among various approaches, data replay has become a promising solution, which can alleviate forgetting by reintroducing representative samples from previous tasks. However, their performance is typically limited by class imbalance, both within the replay buffer due to limited global awareness and between replayed and newly arrived classes. To address this issue, we propose a \underline{c}lass-wise \underline{b}alancing \underline{d}ata \underline{r}eplay method for FCIL (\methodname{}), which employs a global coordination mechanism for class-level memory construction and reweights the learning objective to alleviate the aforementioned imbalances. Specifically, \methodname{} has two key components: 1) the global-perspective data replay module reconstructs global representations of prior task in a privacy-preserving manner, which then guides a class-aware and importance-sensitive sampling strategy to achieve balanced replay; 2) Subsequently, to handle class imbalance across tasks, the task-aware temperature scaling module adaptively adjusts the temperature of logits at both class and instance levels based on task dynamics, which reduces the model’s overconfidence in majority classes while enhancing its sensitivity to minority classes. Experimental results verified that \methodname{} achieves balanced class-wise sampling under heterogeneous data distributions and improves generalization under task imbalance between earlier and recent tasks, yielding a 2\%-15\% Top-1 accuracy improvement over six state-of-the-art methods.

\end{abstract}

\section{Introduction}
Federated learning (FL) is a distributed machine learning paradigm that enables collaborative training of a shared global model across multiple data sources \cite{FedSSA,hu2024fedmut,haozhao1,feng2024federated,fuless}. It periodically performs parameter-level interaction between clients and the server instead of gathering clients' data, which can enhance data privacy while leveraging the diversity of distributed data sources to build a more generalized global model \cite{liao2025privacy,qi2023cross,fu2025learn,hu2024fedcross,wang2024feddse,FedGH}. This mechanism makes it widely applicable to various fields \cite{liang2025tta,feng2025scalable,qi2024attentive,zhong2025unlearning,zhong2022flee}. Building upon this foundation, Federated Class-Incremental Learning (FCIL) extends FL by introducing dynamic data streams where clients sequentially encounter different task classes under non-independent and identically distributed data \cite{dong2022federated,wu2025federated,tran2024text,yang2024federated,fu2025beyond}. However, this amplifies the inherent complexities of FL, as the global model must integrate heterogeneous and evolving knowledge from clients while mitigating catastrophic forgetting, despite having no or only limited access to historical data \cite{lu2024federated,chen2024general,gao2024fedprok}.

\begin{figure}[t]
\centering
\includegraphics[width=1.0\linewidth]{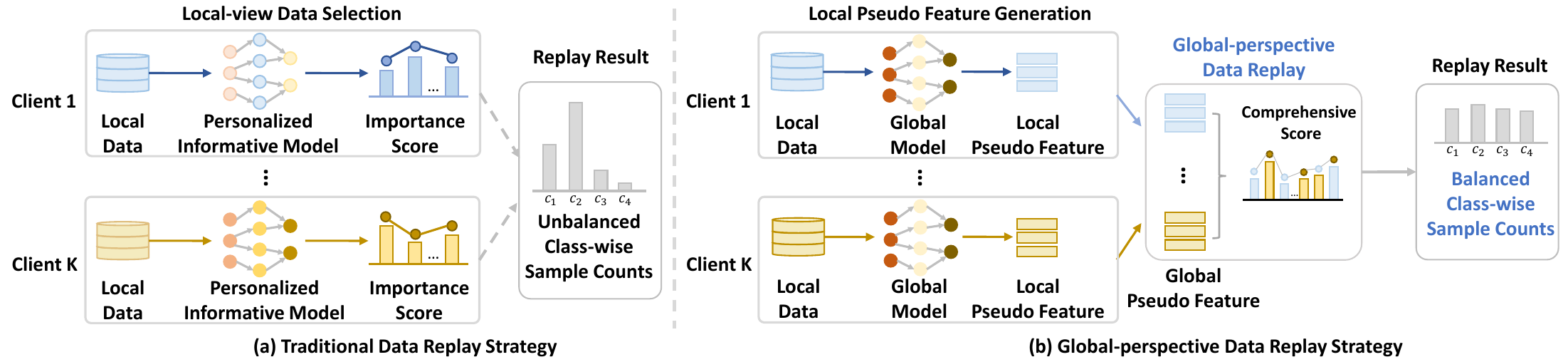}
\caption{Motivation of the \methodname{}. Traditional data replay strategies typically focus on local information and, due to the lack of global awareness, often result in imbalanced class distributions during replay. \methodname{} aims to explore global information in a privacy-preserving manner and leverage it for sampling, which can alleviate the class imbalance problem.
} 
\label{fig1}
\end{figure}

To address the challenge of catastrophic forgetting in FCIL, data replay has emerged as a promising strategy for retaining knowledge from previous tasks. Existing replay-based methods can be broadly categorized into two types: generative-based replay and exemplar-based replay. The former leverages generative models to synthesize representative samples from historical tasks \cite{chen2024general,babakniya2023data,wang2024data}. Its core idea is to learn the data distribution of previous tasks and internalize knowledge in the form of model parameters, enabling the indirect reconstruction of prior knowledge through sample generation when needed \cite{tran2024text,zhang2023target}. However, they often overlook the computational cost of training generative models and are inherently constrained by the quality and fidelity of the synthesized data \cite{chen2024general,liang2024diffusion,yoo2024federated}. In contrast, exemplar-based replay methods directly store real samples from previous tasks, avoiding the complexity of generative processes while leveraging high-quality raw data to ensure robust retention of prior knowledge \cite{dong2022federated,liang2024diffusion,li2024towards,li2025re}. These methods rely on a limited set of historical samples to maintain the decision boundaries of previously learned task classes. However, due to the lack of a global perspective on data distribution across clients, these methods are prone to class-level imbalance in replayed samples, which undermines the model’s ability to retain prior knowledge \cite{li2024towards,li2025re}.



To address these issues, this paper proposes a class-wise balancing data replay method for FCIL, termed \methodname{}, which incorporates the global signal to regulate class-balanced memory construction, aiming to achieve distribution-aware replay and mitigate the challenges posed by non-IID client data, as illustrated in Figure~\ref{fig1}. Specifically, \methodname{} comprises two primary modules: 1) the global-perspective data replay (GDR) module reconstructs a privacy-preserving pseudo global representation of historical tasks by leveraging feature space decomposition, which enables effective cross-client knowledge integration while preserving essential  attributes information. Furthermore, it introduces a principled importance-driven selection mechanism that enables class-balanced replay, guided by a globally-informed understanding of data distribution; 2) the task-aware temperature scaling (TTS) module introduces a multi-level dynamic confidence calibration strategy that combines task-level temperature adjustment with instance-level weighting. By modulating the sharpness of the softmax distribution, it balances the predictive confidence between majority and minority classes, enhancing the model’s robustness to class imbalance between historical and current task samples.

Extensive experiments were conducted on three datasets with different levels of heterogeneity, including performance comparisons, ablation studies, in-depth analysis, and case studies. The results demonstrate that \methodname{} effectively balances the number of  replayed samples across classes and alleviates the long-tail problem. Compared to six state-of-the-art existing methods, \methodname{} achieves a 2\%-15\% Top-1 accuracy improvement.

\section{Related Work}
\subsection{Exemplar-based Replay Methods}
In FCIL, exemplar-based replay methods aim to mitigate catastrophic forgetting by storing and replaying a subset of samples from previous tasks. They typically maintain a small exemplar buffer on each client, which is used during training alongside new task data to preserve knowledge of previously learned classes \cite{dong2022federated,li2024towards,li2025re,rebuffi2017icarl,10323204,li2024sr}. For example, GLFC alleviates forgetting in FCIL by leveraging local exemplar buffers for rehearsal, while introducing class-aware gradient compensation and prototype-guided global coordination to jointly address local and global forgetting \cite{dong2022federated}. Moreover, Re-Fed introduces a Personalized Informative Model to strategically identify and replay task-relevant local samples, enhancing the efficiency of buffer usage and further reducing forgetting in heterogeneous client environments \cite{li2024towards}. However, the lack of global insight in local sample selection often results in class imbalance, while the long-tailed distribution between replayed and current data is frequently overlooked, degrading the effectiveness of data replay \cite{li2024towards,li2025re}.

\subsection{Generative-based Replay Methods}
Generative replay methods aim to reconstruct the samples of past tasks through techniques such as generative modeling \cite{tran2024text,chen2024general,zhang2023target,nguyen2024overcoming,qi2023better}, which enables the model to revisit historical knowledge to mitigate catastrophic forgetting. Following this line of thought, TARGET generates pseudo features through a globally pre-trained encoder and performs knowledge distillation by aligning the current model’s predictions with those of a frozen global model \cite{zhang2023target}; LANDER utilizes pre-trained semantic text embeddings as anchors to synthesize meaningful pseudo samples, and distills knowledge by aligning the model’s predictions with class prototypes derived from textual descriptions \cite{tran2024text}. However, these methods are typically limited by the high computational cost of training generative models and the suboptimal performance caused by low-fidelity pseudo samples \cite{tran2024text,zhang2023target}.

\subsection{Knowledge Distillation-based Methods}
Knowledge distillation-based methods generally follow two paradigms. The first focus om aligning the output predictions of the current model with those of previous models, which aims to preserve task-specific decision boundaries \cite{li2017learning,tan2024fl,xu2025self,psaltis2023fedrcil,chen2025knowledge,zhong2025sacfl,xu2024distribution,dong2023federated_FISS,xu2025long}. The second estimates the importance of model parameters for previously learned tasks and performs regularization to prevent forgetting \cite{kirkpatrick2017overcoming, yu2024overcoming}. Both approaches avoid storing raw data but are prone to knowledge degradation over time, especially as the number of tasks increases \cite{li2017learning,kirkpatrick2017overcoming}.


\section{Preliminaries}
 
We consider a federated class-incremental learning (FCIL) setting, where a central server aims to collaboratively train a global model with the assistance of $K$ distributed clients. Each client $k$ receives a sequence of classification tasks $\{\mathcal{D}_k^{(1)}, \mathcal{D}_k^{(2)}, \dots, \mathcal{D}_k^{(t)}\}$, where each task introduces a disjoint set of new classes. Upon the arrival of task $t$, the global model parameters $\theta_t$ are optimized to minimize the average loss over the union of all samples seen so far, i.e., $\mathbb{D}^t = \bigcup_{s=1}^{t} \bigcup_{k=1}^{K} \mathcal{D}_k^{(s)}$, by solving $\min_{\theta} \frac{1}{|\mathbb{D}^t|} \sum_{s=1}^{t} \sum_{k=1}^{K} \sum_{i=1}^{N_k^{(s)}} \mathcal{L}(f_k(x_{k,i}^{(s)}; \theta),  y_{k,i}^{(s)})$.

In replay-based methods, each client maintains a memory buffer with a fixed budget of $M$ samples. When task $t$ arrives, the client selects up to $N$ representative samples from each of the previous tasks $\{1, \dots, t-1\}$, subject to the total memory constraint. The resulting memory set is denoted by $\mathcal{B}_k^{(t-1)} = \bigcup_{s=1}^{t-1} \{(x_{k,i}^{(s)}, y_{k,i}^{(s)})\}_{i=1}^{N}$, where $N$ is the number of samples stored per task and $\mathcal{B}_k^{(t-1)}$ satisfies $|\mathcal{B}_k^{(t-1)}| \leq M$. The local training set on client $k$ then becomes  $\mathcal{D}_{k,\operatorname{train}}^{(t)} = \mathcal{D}_k^{(t)} \cup \mathcal{B}_k^{(t-1)}$, combining current and replayed samples. Based on these local datasets, the server updates the global model by minimizing the aggregated loss: $\min_{\theta} \sum_{k=1}^K \sum_{(x,y) \in \mathcal{D}_{k,\operatorname{train}}^{(t)}} \mathcal{L}(f_k(x; \theta), y)$.


\section{Class-wise Balancing Data Replay for Federated Class-Incremental Learning}

\begin{figure}[t]
\centering
\includegraphics[width=1.0\linewidth]{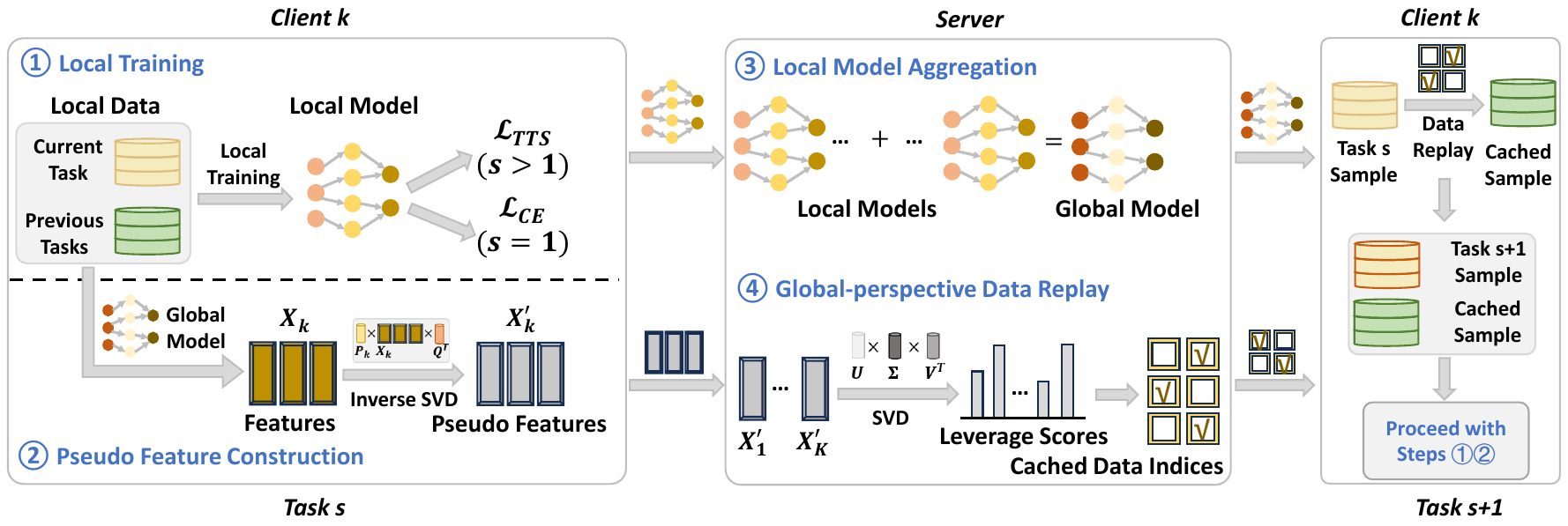}
\caption{Illustration of the \methodname{} framework. It first trains local models using samples from current and previous tasks. After a fixed number of communication rounds, each client extracts local sample features using the global model and applies inverse singular value decomposition (ISVD) to obtain pseudo features. The server then aggregates both local models and pseudo features, performs SVD on the features, and selects representative samples based on leverage scores. The corresponding sample indices are sent back to the clients for balanced replay.
} 
\label{fig2}
\end{figure}

This section presents an effective active data selection method for FCIL, which aims to explore global data distribution to balance class-wise sampling. Moreover, it leverages temperature scaling to adjust the logits, which can alleviate the imbalance between samples from previously learned and newly introduced tasks. Figure \ref{fig2} and Algorithm \ref{alg1} illustrates the framework of the proposed \methodname{}.

\subsection{Global-perspective Data Replay (GDR)}
Due to privacy constraints, traditional data replay strategies typically rely on local data distributions. However, the absence of global information often leads to class imbalance in the replay buffer. To address this, the GDR module aggregates local informative features into a global pseudo feature set, enabling exploration of the global distribution without exposing raw data.

Inspired by Singular Value Decomposition (SVD) \cite{chai2024efficient,an2022numerical}, we first generate a set of random orthogonal matrices: a client-specific matrix $P_k^{(i)} \in \mathbb{R}^{|\mathcal{D}_k^{(i)}| \times |\mathcal{D}_k^{(i)}|}$ for each client $k$ and task $i$, and a globally shared matrix $Q^{(i)} \in \mathbb{R}^{d \times d}$, where $d$ denotes the dimension of the feature. Each client encrypts its local feature matrix $X_k^{(i)} = M_g(\mathcal{D}_k^{(i)})$ via Inverse Singular Value Decomposition (ISVD):
\begin{equation}
    X_k^{(i)'} = P_k^{(i)} X_k^{(i)} Q^{(i)}, \label{eq1}
\end{equation}
and uploads the encrypted matrix $X_k^{(i)'}$ to the server, where $M_g(\cdot)$ is the feature extractor of the global model. The server then aggregates all encrypted matrices into a global matrix $X^{(i)'}$:
\begin{equation}
    X^{(i)'} = \text{concat}\{X_k^{(i)'} \mid k = 1,\ldots,K \}, \label{eq2}
\end{equation}
and performs SVD as follows:
\begin{equation}
    X^{(i)'} = U^{(i)'} \Sigma^{(i)'} V^{(i)'\top}, \label{eq3}
\end{equation}
where $U^{(i)'} \in \mathbb{R}^{n \times n}$, $\Sigma^{(i)'} \in \mathbb{R}^{n \times d}$, and $V^{(i)'} \in \mathbb{R}^{d \times d}$, with $n$ denoting the total number of samples from all clients. Next, the server extracts a submatrix of the left singular vectors for each client $k$ by:
\begin{equation}
    U_k^{(i)} = \mathcal{I}_k(U^{(i)'}) \in \mathbb{R}^{|\mathcal{D}_k^{(i)}| \times n}, \label{eq4}
\end{equation}
where $\mathcal{I}_k(\cdot)$ denotes a row selection function that returns the indices corresponding to client $k$'s samples. To quantify the importance of local samples within the global latent space, client $k$ computes a leverage score \cite{drineas2012fast,pan2016efficient,yu2011dynamic} for $j$-th sample of task $i$ as:
\begin{equation}
    \tau_k^{i,j} = \| e_{i,j}^\top U_k^{(i)} \|_2^2,  \label{eq5}
\end{equation}
where $e_{i,j}$ denotes the $j$-th standard basis vector in task $i$. Notably, a higher leverage score indicates that the sample has a larger projection in the low-dimensional latent space, suggesting that it contributes more significantly to the global structure and is more representative. Moreover, clients send their leverage scores to the server, which aggregates them into a global vector $\tau^i=\text{concat}\{\tau_k^{i,j}|k=1,...,K;j=1,...,n_k^i\}$ and normalizes it to obtain a sampling distribution:
\begin{equation}
    p_k^{i,j} = \frac{\tau_k^{i,j}}{\sum_{j'=1}^{n_k^i} \tau_{k}^{i,j'}}. \label{eq6}
\end{equation}
Subsequently, we perform i.i.d. sampling based on the distribution $\textbf{p}=\{p_k^{i,j}|k=1,...,K;j=1,...,n_k^i\}$. 
Once a sample $x$ is selected, its sampling weight is adjusted to $\frac{1}{\sqrt{n_s \cdot p_x}}e_{x}$, where $n_s$ denotes the number of selected samples and $p_x$ is the original sampling probability of $x$, $e_x$ is the standard basis vector of $x$. This adjustment ensures unbiased estimation during aggregation. Following the sampling procedure, the server communicates the selected sample indices to their respective clients, where the corresponding data points are subsequently marked for further use.

\begin{algorithm}[!t]
\caption{\textsc{FedCBDR}}
\label{alg1}
\begin{algorithmic}[1]

\STATE \textbf{Initialize:} $R$: number of communication rounds; $K$: number of clients; $t$: number of tasks; $\theta_g$: global model parameters; $\mathcal{B}_{k}^{pre}$: replay buffer for historical tasks on client $k$; $\mathcal{D}_{k}^{s}$: local data of task $s$ on client $k$.

\FOR {each task $s = 1$ to $t$}
    \FOR {each communication round $r = 1$ to $R$}
        \FOR {each client $k = 1$ to $K$}
            \STATE Initialize local model parameters: $\theta_k \leftarrow \theta_g$
            \IF{$s == 1$} 
                \STATE Sample a mini-batch $\zeta$ from $\mathcal{D}_k^{(1)}$, and update $\theta_k$ using Eq.~\ref{eq8}.
            \ELSE
                \STATE Store the historical task data corresponding to globally sampled IDs into $\mathcal{B}_{k}^{pre}$.
                \STATE Sample a mini-batch $\zeta$ from $\mathcal{D}_k^{(s)} \cup \mathcal{B}_{k}^{pre}$, and update $\theta_k$ using Eq.~\ref{eq9}.
                \STATE Compute pseudo-features based on Eq.~\ref{eq1}, and upload them to the server.
            \ENDIF
        \ENDFOR
        
        \IF{$r < R$}
            \STATE Aggregate local model parameters across clients.
        \ELSE
            \STATE Aggregate model parameters and pseudo-features from all clients using Eq.~\ref{eq2}.
            \STATE Perform \textbf{Global Sampling} based on Eqs.~\ref{eq3}--\ref{eq6}, and send the selected sample IDs back to the corresponding clients.
        \ENDIF
    \ENDFOR
\ENDFOR

\vspace{1mm}
\STATE \textbf{// Global Sampling Procedure}
\STATE Form the global feature pool $X^{(i)}$ by aggregating all pseudo-features via Eq.~\eqref{eq2}.
\STATE Perform singular value decomposition (SVD) using Eq.~\ref{eq3} to extract key attributes.
\STATE Compute leverage scores for each client’s samples using Eqs.~\ref{eq4}--\ref{eq5}, and normalize globally using Eq.~\ref{eq6}.
\STATE Perform sampling and adjust the probabilities of the selected samples accordingly.
\end{algorithmic}
\end{algorithm}

\subsection{Task-aware Temperature Scaling (TTS)}
Due to limited replay budgets, samples from previous tasks are often much fewer than those from the current task, leading to class imbalance and poor retention of past knowledge. To mitigate this, the TTS module dynamically adjusts sample temperature and weight based on task order, enhancing the contribution of tail-class samples during optimization.

Specifically, we use a lower temperature to sharpen logits for samples from earlier tasks. Furthermore, to further amplify the optimization effect of tail-class samples during training, we also leverage a re-weighted cross-entropy loss, i.e.,
\begin{equation}
\resizebox{0.94\hsize}{!}{$
\mathcal{L}_{\text{TTS}} = \frac{1}{N_{\text{old}}} \displaystyle\sum_{i=1}^{N_{\text{old}}} \omega_{\text{old}} \cdot \mathrm{CE} \left( y_i, \mathrm{Softmax} \left( \mathrm{Concat} \left( \frac{z_i^{\text{old}}}{\tau_{\text{old}}}, \frac{z_i^{\text{new}}}{\tau_{\text{new}}} \right) \right) \right) + \frac{1}{N_{\text{new}}} \displaystyle\sum_{j=1}^{N_{\text{new}}} \omega_{\text{new}} \cdot \mathrm{CE} \left( y_j, \mathrm{Softmax} \left( \mathrm{Concat} \left( \frac{z_j^{\text{old}}}{\tau_{\text{old}}}, \frac{z_j^{\text{new}}}{\tau_{\text{new}}} \right) \right) \right)
$}
\end{equation}
where $N_{\text{old}}$ and $N_{\text{new}}$ denote the number of samples from the previous and newly arrived task, respectively; $y_i$ and $y_j$ are the ground-truth labels; $z_i^{\text{old}}$ and $z_i^{\text{new}}$ denote the logits corresponding to old classes and new classes, respectively; $\tau_{\text{old}}$ and $\tau_{\text{new}}$ are the temperature scaling factors for previous and newly arrived task samples; $\omega_{\text{old}}$ and $\omega_{\text{new}}$ are the corresponding sample weights; $\mathrm{CE}(\cdot)$ denotes the cross-entropy loss function; and $\mathrm{Softmax}(z/\tau)$ is the temperature-scaled softmax function used to adjust the sharpness of the output distribution.

\subsection{Training Strategy}
The training strategy consists of two stages to progressively address the evolving challenges in federated class-incremental learning. Algorithm \ref{alg1} presents the pipeline of the \methodname{}.

\textbf{Stage 1: Initial Task Optimization.}  
In the first task, client $k$ learns from local data using the standard cross-entropy loss, i.e.,
\begin{equation}
\min_{\theta_k} \; \frac{1}{N} {\textstyle \sum_{i=1}^{N}}  \mathrm{CE}(y_i, \mathrm{Softmax}(f_{\theta_k}(x_i))),
\label{eq8}
\end{equation}

\textbf{Stage 2: Class-Incremental Optimization.}                   
As new tasks arrive and class imbalance emerges between previous and current tasks in client $k$, we employ $\mathcal{L}_{TTS}$ to mitigate the imbalance, i.e.,
\begin{equation}
\resizebox{0.94\hsize}{!}{$
\min_{\theta_k} \; 
\frac{1}{N_{\text{old}}} \displaystyle\sum_{i=1}^{N_{\text{old}}} \omega_{\text{old}} \cdot \mathrm{CE}\left(y_i, \mathrm{Softmax}\left(\mathrm{Concat}\left(\frac{f_{\theta_k}^{\text{old}}(x_i)}{\tau_{\text{old}}}, \frac{f_{\theta_k}^{\text{new}}(x_i)}{\tau_{\text{new}}} \right)\right)\right)
+
\frac{1}{N_{\text{new}}} \displaystyle\sum_{j=1}^{N_{\text{new}}} \omega_{\text{new}} \cdot \mathrm{CE}\left(y_j, \mathrm{Softmax}\left(\mathrm{Concat}\left(\frac{f_{\theta_k}^{\text{old}}(x_j)}{\tau_{\text{old}}}, \frac{f_{\theta_k}^{\text{new}}(x_j)}{\tau_{\text{new}}} \right)\right)\right)
$}
\label{eq9}
\end{equation}
where $x_i$ is the input sample, $y_i$ is the corresponding ground-truth, $f_{\theta_k}^{\text{old}}(x)$ and $f_{\theta_k}^{\text{new}}(x)$ represent the outputs of the model corresponding to old and new classes, respectively. $\mathrm{Softmax}(\cdot)$ converts the logits into a probability distribution.

\section{Experiments}
\begin{table}[t]
\centering
\caption{Statistics of the datasets used in experiments.}
\setlength{\tabcolsep}{3.6pt}
\begin{tabular}{l|c|c|c|c|ccc}
\toprule
\multirow{2}{*}{\textbf{Datasets}} 
& \multirow{2}{*}{\textbf{\#Class}} 
& \multirow{2}{*}{\textbf{\#Training}} 
& \multirow{2}{*}{\textbf{\#Testing}} 
& \multirow{2}{*}{\textbf{Image Size}} 
& \multicolumn{3}{c}{\textbf{Federated Settings}} \\
\cmidrule(lr){6-8}
& & & & & \textbf{Clients} & \textbf{Tasks} & \textbf{Heterogeneity} \\
\midrule
CIFAR10       & 10  & 50,000  & 10,000  & 32 $\times$ 32 & 5/10 & 3/5 & 0.5/1.0 \\
CIFAR100      & 100 & 50,000  & 10,000  & 32 $\times$ 32 & 5/10 & 5/10 & 0.1/0.5/1.0 \\
TinyImageNet  & 200 & 100,000 & 10,000  & 64 $\times$ 64 & 5/10 & 10/20 & 0.1/0.5/1.0 \\
\bottomrule
\end{tabular}
\label{tab5}
\end{table}
\subsection{Experiment Settings}
\paragraph{Datasets.}
Following existing studies \cite{zhang2023target,li2024towards}, we conducted all experiments on three commonly used datasets, including CIFAR10 \cite{krizhevsky2009learning,qi2022clustering}, CIFAR100 \cite{krizhevsky2009learning,qi2022clustering} and TinyImageNet \cite{le2015tiny} to validate the effectiveness of the \methodname{}. We simulate heterogeneous data distributions across clients using the Dirichlet distribution with parameters $\beta = \{0.1,0.5,1.0\}$, where smaller values of $\beta$ correspond to higher level of data heterogeneity. The statistical details are presented in the Table \ref{tab5}.

\paragraph{Evaluation Metric.}
Following prior studies \cite{tran2024text,chen2024flexfl,wang2024fednlr,qi2024cross}, we adopt Top-1 Accuracy as the evaluation metric, defined as $\text{Accuracy} = N_{\text{correct}} / N_{\text{total}}$, where $N_{\text{correct}}$ and $N_{\text{total}}$ denote the number of correct predictions and the total number of samples, respectively.

\paragraph{Implementation Details.} \label{sec513}
In the experiments, the number of clients is fixed at $K=5$, with each client running local epochs $E=2$ per round, using a batch size $B=128$. For all datasets, we adopt ResNet-18 as the backbone, with the classifier's output dimension dynamically updated as tasks progress and conduct $T=100$ communication rounds per task. The SGD optimizer is employed with a learning rate of $0.01$ and a weight decay of $1 \times 10^{-5}$. The number of stored samples per task varies by dataset and split setting: for CIFAR10, 450 samples are stored under 3-task splits and 300 under 5-task splits; for CIFAR100, 1,000 samples are used for 5-task splits and 500 for 10-task splits; for TinyImageNet, 2,000 samples are stored for 10-task splits and 1,000 for 20-task splits. For the temperature and weighted parameters, we select $\tau_{old} \in \{0.8, 0.9\}$ and $w_{old} \in \{1.1, 1.2, 1.3, 1.4\}$ for previous tasks, while $\tau_{new} \in \{1.1, 1.2\}$ and $w_{new} \in \{0.7, 0.8, 0.9\}$ are used for newly arrived tasks. Moreover, the hyperparameters of baselines are tuned based on their original papers for fair comparison. And training on each client is performed using an NVIDIA RTX 3090 GPU (24 GB).

\begin{table*}[t]
\centering
\caption{Performance comparison between \methodname{} and baselines across three datasets under varying levels of heterogeneity ($\beta$). CIFAR10 is divided into 3 tasks, CIFAR100 into 5 tasks, and TinyImageNet into 10 tasks. All methods were executed under three different random seeds, and both the mean and standard deviation of the results are reported. The best results are \textbf{bolded}.}
\renewcommand{\arraystretch}{1.}
\setlength{\tabcolsep}{2.pt}
\begin{tabular}{c|cc|ccc|ccc}
\toprule
\multirow{2}{*}{\textbf{Method}} & \multicolumn{2}{c|}{\textbf{CIFAR10}} & \multicolumn{3}{c|}{\textbf{CIFAR100}} & \multicolumn{3}{c}{\textbf{TinyImageNet}} \\ \cline{2-9}
& $\beta{=}0.5$ & $\beta{=}1.0$ & $\beta{=}0.1$ & $\beta{=}0.5$ & $\beta{=}1.0$ & $\beta{=}0.1$ & $\beta{=}0.5$ & $\beta{=}1.0$  \\
\midrule
Finetune & 38.71$_{\pm 3.7}$ & 40.49$_{\pm 3.0}$ & 15.17$_{\pm 2.2}$ & 16.75$_{\pm 2.6}$ & 17.15$_{\pm 1.3}$ & 6.06$_{\pm 0.9}$ & 6.00$_{\pm 0.8}$ & 6.40$_{\pm 0.5}$ \\
FedEWC  & 39.93$_{\pm 1.1}$ & 42.70$_{\pm 2.5}$ & 18.30$_{\pm 2.4}$ & 20.70$_{\pm 5.3}$ & 21.22$_{\pm 3.4}$ & 6.30$_{\pm 0.8}$ & 6.94$_{\pm 0.7}$ & 7.36$_{\pm 0.6}$ \\
FedLwF & 56.03$_{\pm 1.6}$ & 58.29$_{\pm 3.6}$ & 33.97$_{\pm 2.6}$ & 37.09$_{\pm 3.1}$ & 41.91$_{\pm 2.5}$ & 11.81$_{\pm 0.9}$ & 11.47$_{\pm 1.0}$ & 14.87$_{\pm 1.2}$ \\
TARGET & 44.17$_{\pm 4.4}$ & 54.49$_{\pm 4.5}$ & 30.15$_{\pm 3.6}$ & 33.47$_{\pm 4.3}$ & 35.25$_{\pm 2.0}$ & 10.71$_{\pm 1.4}$ & 10.18$_{\pm 0.9}$ & 12.49$_{\pm 1.1}$ \\
LANDER & 53.90$_{\pm 3.2}$ & 60.79$_{\pm 1.4}$ & 44.07$_{\pm 3.3}$ & 47.63$_{\pm 3.7}$ & \textbf{52.77}$_{\pm 1.4}$ & 13.80$_{\pm 0.8}$ & 15.02$_{\pm 1.9}$ & 16.36$_{\pm 1.0}$ \\
Re-Fed  & 53.46$_{\pm 3.5}$ & 60.73$_{\pm 4.3}$ & 32.67$_{\pm 3.7}$ & 38.42$_{\pm 2.9}$ & 45.28$_{\pm 2.6}$ & 15.73$_{\pm 1.7}$ & 15.93$_{\pm 1.3}$ & 16.05$_{\pm 1.1}$ \\\hline
\textbf{\methodname{}} & \textbf{64.11}$_{\pm 1.2}$ & \textbf{65.20}$_{\pm 1.9}$ & \textbf{46.40}$_{\pm 1.6}$ & \textbf{49.76}$_{\pm 2.7}$ & 52.06$_{\pm 1.5}$ & \textbf{18.37}$_{\pm 1.1}$ & \textbf{18.86}$_{\pm 0.9}$ & \textbf{18.78}$_{\pm 0.9}$ \\
\bottomrule
\end{tabular}
\label{tab1}
\end{table*}

\begin{table*}[t]
\centering
\caption{Performance comparison between \methodname{} and baselines across three datasets under varying levels of heterogeneity  ($\beta$). CIFAR10 is divided into 5 tasks, CIFAR100 into 10 tasks, and TinyImageNet into 20 tasks. All methods were executed under three different random seeds, and both the mean and standard deviation of the results are reported. The best results are \textbf{bolded}.}
\renewcommand{\arraystretch}{1.}
\setlength{\tabcolsep}{2.pt}
\begin{tabular}{c|cc|ccc|ccc}
\toprule
\multirow{2}{*}{\textbf{Method}} & \multicolumn{2}{c|}{\textbf{CIFAR10}} & \multicolumn{3}{c|}{\textbf{CIFAR100}} & \multicolumn{3}{c}{\textbf{TinyImageNet}} \\\cline{2-9}
& $\beta{=}0.5$ & $\beta{=}1.0$ & $\beta{=}0.1$ & $\beta{=}0.5$ & $\beta{=}1.0$ & $\beta{=}0.1$ & $\beta{=}0.5$ & $\beta{=}1.0$  \\
\midrule
Finetune & 19.78$_{\pm 2.3}$ & 23.34$_{\pm 2.8}$ & 7.22$_{\pm 1.1}$ & 9.39$_{\pm 0.7}$ & 9.64$_{\pm 0.5}$ & 3.40$_{\pm 0.4}$ & 3.73$_{\pm 0.5}$ & 3.95$_{\pm 0.3}$ \\
FedEWC & 20.11$_{\pm 2.7}$ & 28.97$_{\pm 2.3}$ & 8.08$_{\pm 0.3}$ & 11.69$_{\pm 0.7}$ & 12.19$_{\pm 1.7}$ & 3.50$_{\pm 0.3}$ & 4.58$_{\pm 0.4}$ & 5.08$_{\pm 0.9}$ \\
FedLwF  & 38.76$_{\pm 2.3}$ & 52.95$_{\pm 3.1}$ & 18.73$_{\pm 1.1}$ & 25.30$_{\pm 0.6}$ & 28.21$_{\pm 1.0}$ & 3.67$_{\pm 0.4}$ & 6.61$_{\pm 0.6}$ & 10.22$_{\pm 1.3}$ \\
TARGET & 35.27$_{\pm 1.7}$ & 48.28$_{\pm 1.2}$ & 13.61$_{\pm 0.8}$ & 21.09$_{\pm 0.4}$ & 24.22$_{\pm 1.1}$ & 5.32$_{\pm 0.6}$ & 5.39$_{\pm 0.6}$ & 5.72$_{\pm 0.5}$ \\
LANDER & 40.22$_{\pm 2.4}$ & 58.07$_{\pm 3.4}$ & 27.79$_{\pm 1.9}$ & 33.51$_{\pm 2.3}$ & 37.42$_{\pm 1.8}$ & 8.89$_{\pm 0.6}$ & 8.57$_{\pm 0.8}$ & 10.45$_{\pm 0.6}$ \\
Re-Fed & 54.94$_{\pm 3.1}$ & 58.19$_{\pm 2.5}$ & 29.33$_{\pm 1.3}$ & 39.54$_{\pm 1.3}$ & 40.96$_{\pm 1.1}$ & 9.36$_{\pm 0.9}$ & 11.44$_{\pm 0.7}$ & 12.27$_{\pm 1.1}$ \\\hline
\textbf{\methodname{}} & \textbf{61.18}$_{\pm 1.3}$ & \textbf{65.42}$_{\pm 1.8}$ & \textbf{45.11}$_{\pm 1.2}$ & \textbf{46.51}$_{\pm 1.6}$ & \textbf{47.79}$_{\pm 1.4}$ & \textbf{12.58}$_{\pm 0.4}$ & \textbf{14.47}$_{\pm 0.7}$ & \textbf{15.69}$_{\pm 0.6}$ \\
\bottomrule
\end{tabular}
\label{tab2}
\end{table*}

\subsection{Performance Comparison}
To evaluate the effectiveness of the proposed \methodname{}, we compare it with six representative baseline methods: Finetune \cite{tran2024text}, FedEWC \cite{kirkpatrick2017overcoming}, FedLwF \cite{li2017learning}, TARGET \cite{zhang2023target}, LANDER \cite{tran2024text}, and Re-Fed \cite{li2024towards}.
As reported in Table~\ref{tab1} and Table~\ref{tab2}, the results can be summarized as follows:

\begin{itemize}[leftmargin=10pt]
\item \methodname{} achieves the highest Top-1 accuracy in most cases across the three datasets under varying levels of heterogeneity and task splits. The only suboptimal result occurs on CIFAR100 with 5 tasks and $\beta=1.0$, where \methodname{} (52.06\%) performs slightly worse than LANDER (52.77\%). This demonstrates the adaptability and robustness of the proposed \methodname{} across complex settings.
\item Despite LANDER attains the best performance on CIFAR100 under the 5-task and $\beta=1.0$ setting, it demands the generation of more than 10,000 samples per task, and the overhead of training its data generator surpasses that of the federated model, raising concerns about its scalability.
\item Knowledge distillation-based methods like FedLwF perform well on simpler tasks (CIFAR10) by using pretrained knowledge to guide local models. However, their performance drops on more complex or heterogeneous tasks due to limited adaptability to local variations.
\item Given an equal memory budget, class-balanced sampling (\methodname{}) consistently achieves superior performance compared to class-imbalanced strategy (Re-Fed), as it ensures more equitable representation across categories and effectively mitigates class-level forgetting in FCIL scenarios.
\end{itemize}

\subsection{Ablation Study}
In this section, we conducted an ablation study to investigate the contributions of key modules, including the  Global-perspective Active Data Replay (GDR) module and the Task-aware Temperature Scaling (TTS) module. Table~\ref{tab3} presents the results, which can be summarized as follows:

\begin{itemize}[leftmargin=10pt]
\item Incorporating the GDR module  substantially improves performance across all cases, particularly under high data heterogeneity ($\beta=0.1$), demonstrating its effectiveness in alleviating catastrophic forgetting even with a limited number of replay samples in federated class-incremental learning.

\item Using the TTS module alone leads to consistent improvements over Finetune, highlighting its effectiveness in addressing intra-client class imbalance through temperature scaling. This contribution to better generalization is particularly evident under the more challenging "5/10/20" task splitting scenario.

\item The integration of both modules results in the best overall performance, consistently achieving the highest Top-1 accuracy across various datasets and heterogeneity levels. This stems from their complementary strengths: the GDR module mitigates inter-task forgetting, while the TTS module alleviates both intra- and inter-client class imbalance.
\end{itemize}

\begin{table*}[t]
\centering
\caption{Ablation results under different levels of data heterogeneity and task splitting settings. “3/5/10” denotes CIFAR10 with 3 tasks, CIFAR100 with 5 tasks, and TinyImageNet with 10 tasks; “5/10/20” represents 5, 10, and 20 tasks respectively.}
\renewcommand{\arraystretch}{1.05}
\setlength{\tabcolsep}{2pt} 
\resizebox{\textwidth}{!}{
\begin{tabular}{c|c|cc|ccc|ccc}
\toprule
\multirow{2}{*}{\makecell[c]{\textbf{Task} \\ \textbf{Splitting}}} & \multirow{2}{*}{\textbf{Method}} & \multicolumn{2}{c|}{\textbf{CIFAR10}} & \multicolumn{3}{c|}{\textbf{CIFAR100}} & \multicolumn{3}{c}{\textbf{TinyImageNet}} \\\cline{3-10}
& & $\beta{=}0.5$ & $\beta{=}1.0$ & $\beta{=}0.1$ & $\beta{=}0.5$ & $\beta{=}1.0$ & $\beta{=}0.1$ & $\beta{=}0.5$ & $\beta{=}1.0$ \\
\midrule
\multirow{4}{*}{3/5/10} 
& Finetune & 38.71$_{\pm 3.7}$ & 40.49$_{\pm 3.0}$ & 15.17$_{\pm 2.2}$ & 16.75$_{\pm 2.6}$ & 17.15$_{\pm 1.3}$ & 6.06$_{\pm 0.9}$ & 6.00$_{\pm 0.8}$ & 6.40$_{\pm 0.5}$ \\
& +GDR & 62.13$_{\pm 2.1}$ & 63.81$_{\pm 1.9}$ & 45.28$_{\pm 1.5}$ & 47.66$_{\pm 0.9}$ & 51.47$_{\pm 1.7}$ & 17.24$_{\pm 0.6}$ & 17.89$_{\pm 0.5}$ & 18.04$_{\pm 0.4}$ \\
& +TTS & 41.34$_{\pm 2.3}$ & 42.55$_{\pm 2.2}$ & 17.32$_{\pm 0.5}$ & 17.14$_{\pm 0.4}$ & 19.32$_{\pm 0.5}$ & 6.67$_{\pm 0.2}$ & 6.92$_{\pm 0.3}$ & 7.27$_{\pm 0.4}$ \\
& +GDR+TTS & \textbf{64.11}$_{\pm 1.2}$ & \textbf{65.20}$_{\pm 1.9}$ & \textbf{46.40}$_{\pm 1.6}$ & \textbf{49.76}$_{\pm 2.7}$ & \textbf{52.06}$_{\pm 1.5}$ & \textbf{18.37}$_{\pm 1.1}$ & \textbf{18.86}$_{\pm 0.9}$ & \textbf{18.78}$_{\pm 0.9}$  \\
\midrule
\multirow{4}{*}{5/10/20} 
& Finetune & 19.78$_{\pm 2.3}$ & 23.34$_{\pm 2.8}$ & 7.22$_{\pm 1.1}$ & 9.39$_{\pm 0.7}$ & 9.64$_{\pm 0.5}$ & 3.40$_{\pm 0.4}$ & 3.73$_{\pm 0.5}$ & 3.95$_{\pm 0.3}$ \\
& +GDR & 59.34$_{\pm 3.1}$ & 63.20$_{\pm 2.6}$ & 44.04$_{\pm 1.3}$ & 46.33$_{\pm 0.5}$ & 46.50$_{\pm 0.8}$ & 11.44$_{\pm 0.3}$ & 13.85$_{\pm 0.5}$ & 14.51$_{\pm 0.6}$ \\
& +TTS & 22.43$_{\pm 2.4}$ & 25.81$_{\pm 2.1}$ & 8.31$_{\pm 0.2}$ & 10.21$_{\pm 0.3}$ & 10.33$_{\pm 0.4}$ & 3.78$_{\pm 0.5}$ & 4.04$_{\pm 0.4}$ & 4.16$_{\pm 0.3}$ \\
& +GDR+TTS & \textbf{61.18}$_{\pm 1.3}$ & \textbf{65.42}$_{\pm 1.8}$ & \textbf{45.11}$_{\pm 1.2}$ & \textbf{46.51}$_{\pm 1.6}$ & \textbf{47.79}$_{\pm 1.4}$ & \textbf{12.58}$_{\pm 0.4}$ & \textbf{14.47}$_{\pm 0.7}$ & \textbf{15.69}$_{\pm 0.6}$ \\
\bottomrule
\end{tabular}
}
\label{tab3}
\end{table*}

\subsection{Performance Evaluation of \methodname{} under Incremental Tasks}

\begin{figure}[h]
\centering
\includegraphics[width=1.0\linewidth]{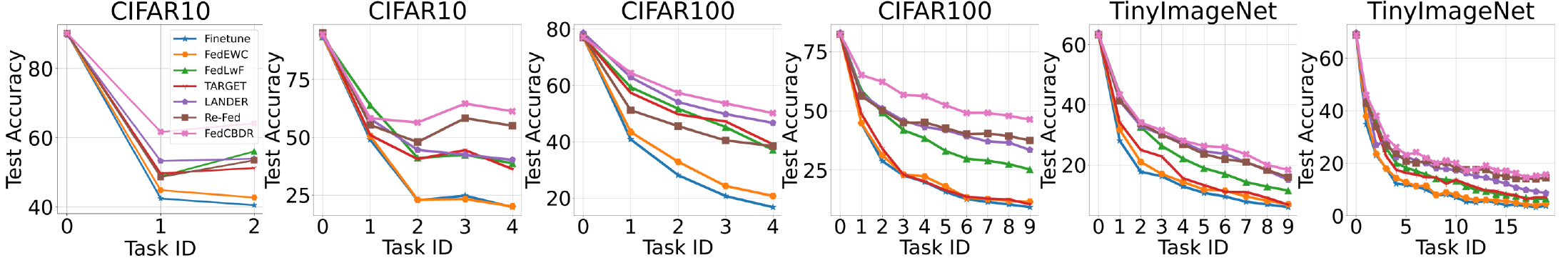}
\caption{Performance comparison of all methods across varying task splits on CIFAR10 (3/5 tasks), CIFAR100 (5/10 tasks), and TinyImageNet (10/20 tasks) with $\beta=0.5$.
} 
\label{fig3}
\end{figure}

This section investigates the performance of \methodname{} and the baselines in incremental cases on three datasets. Figure \ref{fig3} presents the average accuracy of all methods on both current and previous tasks. Notably, \methodname{} consistently outperforms other baseline methods across all task splits, with its accuracy curves remaining higher throughout the incremental process. Furthermore, \methodname{} exhibits a slower performance degradation as the number of tasks increases, indicating stronger resistance to catastrophic forgetting. In addition, it maintains significantly higher accuracy on later tasks, especially in challenging settings such as CIFAR100 and TinyImageNet with 10 tasks, highlighting its ability to balance knowledge retention and adaptation to new classes.

\subsection{Quantitative Analysis of Replay Buffer Size on Test Accuracy}
\begin{table}[h]
\centering
\caption{Comparison of model performance with varying memory size $M$ across datasets.}
\renewcommand{\arraystretch}{1}
\setlength{\tabcolsep}{3.2pt}
\begin{tabular}{c|ccc|ccc|cc}
\toprule
\multirow{2}{*}{\textbf{Methods}} & \multicolumn{3}{c|}{\textbf{CIFAR10}} & \multicolumn{3}{c|}{\textbf{CIFAR100}} & \multicolumn{2}{c}{\textbf{TinyImageNet}} \\\cline{2-9}
 & M=150 & M=300 & M=450 & M=500 & M=1000 & M=1500 & M=2000 & M=2500 \\
\midrule
LANDER (10240) & \multicolumn{3}{c|}{52.90} & \multicolumn{3}{c|}{47.05}  & \multicolumn{2}{c}{14.77} \\
Re-Fed & 47.23 & 53.47 & 54.66 & 33.89 & 38.42 & 47.84 & 15.89 & 16.78  \\
\methodname{} & 51.99 & 59.02 & 63.81 & 40.12 & 49.66 & 55.94 & 18.33 & 19.41 \\
\bottomrule
\end{tabular}
\label{tab4}
\end{table}
In this section, we evaluate the performance of Re-Fed and \methodname{} under different buffer size $M$ settings, and additionally include LANDER, which generates 10,240 synthetic samples for each task. As shown in Table \ref{tab4}, \methodname{} exhibits more significant performance advantages over Re-Fed under limited memory settings, and even surpasses LANDER, which relies on a large-scale generative replay buffer. Furthermore, as the buffer size increases, \methodname{} demonstrates more stable and significant performance improvements. This indicates that the method can effectively leverage larger replay buffers for continuous optimization. However, Re-Fed exhibits noticeable performance fluctuations under small and medium buffer settings. In particular, its accuracy is significantly lower than that of \methodname{} on CIFAR100 with 
$M=500$ and TinyImageNet with 
$M=2000$, indicating its limited ability to mitigate inter-class interference and retain knowledge from previous tasks. These findings validate that, under the same buffer budget, a balanced sampling distribution is more effective than an imbalanced one in alleviating forgetting and improving overall model performance.

\subsection{Sensitivity Analysis of \methodname{} on Temperature and Weighted Hyperparameters}
\begin{figure}[h]
\centering
\includegraphics[width=1.0\linewidth]{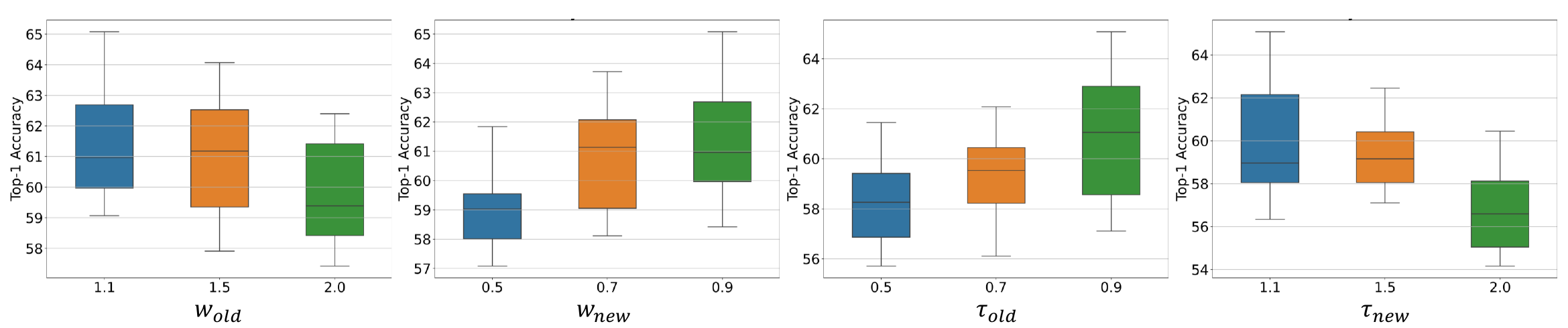}
\caption{Performance of \methodname{} on CIFAR100 ($\beta=0.5$, 5-task split) under varying temperature ($\tau_{old}\in\{0.5,0.7,0.9\}$, $\tau_{new}\in\{1.1,1.5,2.0\}$) and weighted ($w_{old}\in\{1.1,1.5,2.0\}$, $w_{new}\in\{0.5,0.7,0.9\}$) settings.
} 
\label{fig4}
\end{figure}
Figure \ref{fig4} gives a sensitivity analysis of \methodname{} with respect to temperature and sample weighting hyperparameters. Overall, temperature scaling and sample re-weighting help mitigate class imbalance, but model performance varies considerably with different hyperparameter settings. The model achieves better overall performance when $\omega_{\text{old}} = 1.1$, $\omega_{\text{new}} = 0.9$, $\tau_{\text{old}} = 0.9$, and $\tau_{\text{new}} = 1.1$. This is because slightly higher weight and temperature for previous-task samples help retain old knowledge, while lower weight and higher temperature for newly arrived samples reduce overfitting and improve adaptation.
However, inappropriate hyperparameter choices may harm performance. For instance, a large $\tau_{\text{new}}$ (e.g., 2.0) leads to overly smooth predictions, reducing discrimination among newly arrived classes. These results emphasize the need for proper tuning to ensure balanced learning.

\subsection{Comparison of Per-Class Sample Distributions in the Replay Buffer}
\begin{figure}[h]
\centering
\includegraphics[width=1.0\linewidth]{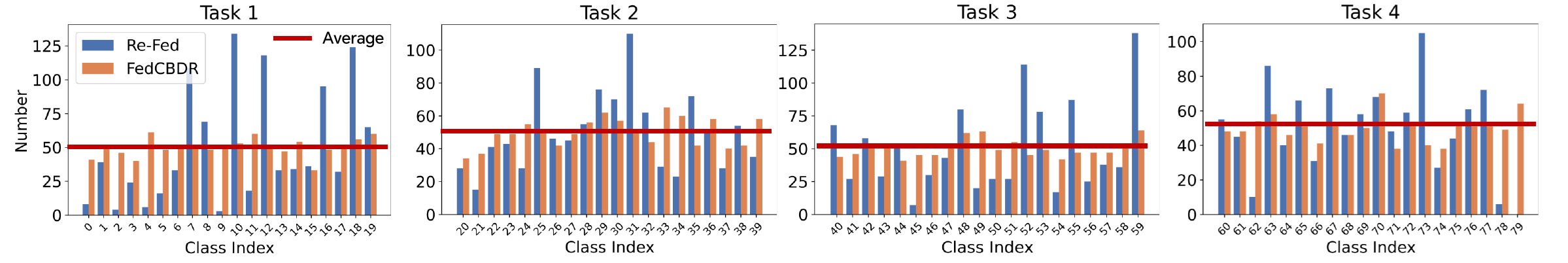}
\caption{Comparison of per-class sample distributions in the replay buffer between \methodname{} and Re-Fed on the CIFAR100 dataset, under a heterogeneity level of 
$\beta=0.5$ and a 5-task split case.
} 
\label{fig5}
\end{figure}
To evaluate the effectiveness of \methodname{} in balancing class-wise sampling, Figure \ref{fig5} illustrates the per-class sample distributions in the replay buffer between \methodname{} and Re-Fed across different task stages. Overall, across different task stages, \methodname{} (orange bars) exhibits a per-class sample distribution that is consistently closer to the average level (red line), whereas Re-Fed shows noticeable skewness and fluctuations. This indicates that \methodname{} is more effective in achieving balanced class-wise sampling in the replay buffer. In addition, \methodname{} ensures that no class is overlooked during sampling, while Re-Fed may fail to retain certain classes in the replay buffer—for example, class 79 is missing in Task 4 under Re-Fed. This highlights the robustness of \methodname{} in maintaining class coverage throughout incremental learning.

\subsection{Visualization of Model Attention and Temperature-aware Logits Adjustment}
\begin{figure}[h]
\centering
\includegraphics[width=1.0\linewidth]{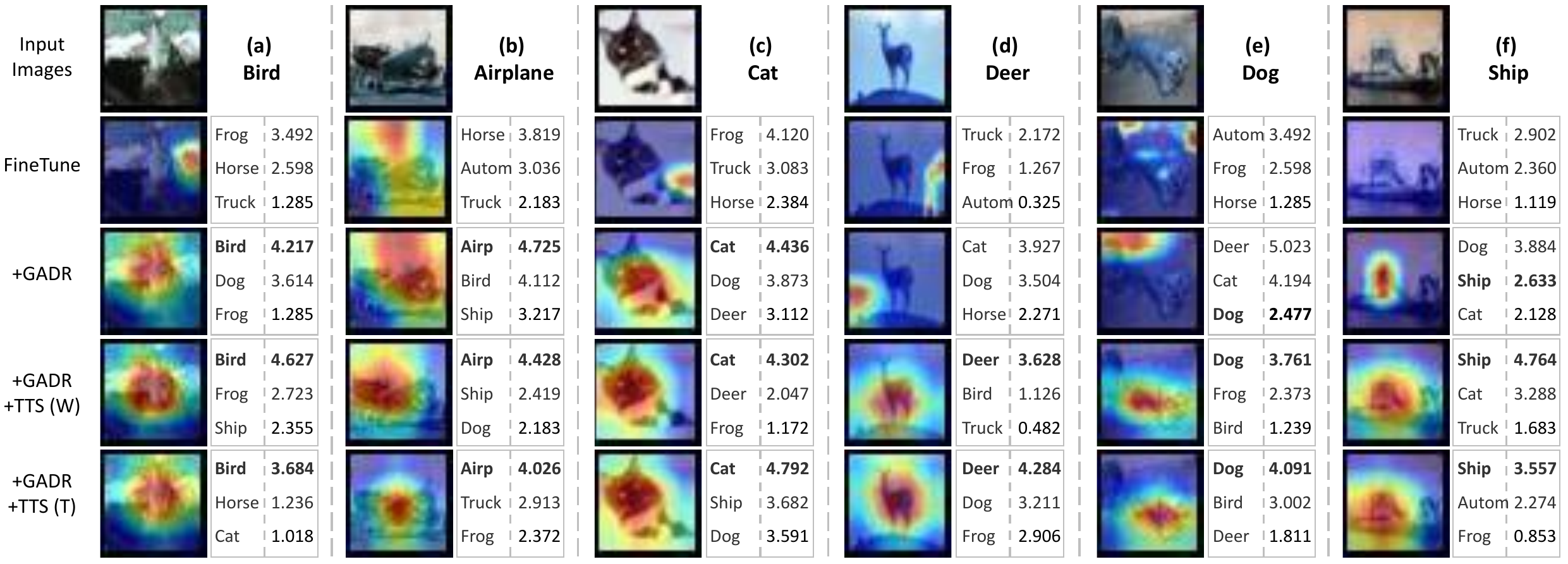}
\caption{Case studies of model attention and the effect of temperature-aware logits adjustment on CIFAR10 ($\beta=0.5$, 3-task split).}
\label{fig6}
\end{figure}
This section presents case studies comparing prediction confidence and attention focus using Grad-CAM \cite{qi2025cross,10605121,qi2025federated,meng2025causal,wbk_mm2017} visualizations. As shown in Figure \ref{fig6}(a-c), in the absence of data replay, the model struggles to correctly classify samples from previous tasks and fails to attend to the relevant target regions. The incorporation of data replay in \methodname{} alleviates this issue by correcting predictions and guiding attention back to semantically important areas. Despite partially mitigating forgetting, data replay alone may still lead to misclassification or low-confidence predictions for tail classes with limited samples. The integration of temperature scaling (T) and sample re-weighting (W) in the \texttt{TTS} module enables the model to better distinguish confusing classes through temperature adjustment, improving tail class accuracy and enhancing prediction stability, as depicted in Figure \ref{fig6}(d-f). These findings demonstrate the crucial role of the collaboration between both modules in mitigating knowledge forgetting during incremental learning.

\section{Conclusions and Future Work}
To address the challenge of inter-class imbalance in replay-based federated class-incremental learning, we propose \methodname{} that combines class-balanced sampling with loss adjustment to better exploit the global data distribution and enhance the contribution of tail-class samples to model optimization. Specifically, it uses SVD to decouple and reconstruct local data, aggregates local information in a privacy-preserving manner, and explores i.i.d. sampling within the aggregated distribution. In addition, it applies task-aware temperature scaling and sample re-weighting to mitigate the long-tail problem. Experimental results show that \methodname{} effectively reduces inter-class sampling imbalance and significantly improves final performance.

Despite the impressive performance of \methodname{}, there remain several directions worth exploring to address its limitations. Specifically, we plan to investigate lightweight sampling strategies to reduce feature transmission costs in \methodname{}, and to develop more robust post-sampling balancing methods that mitigate class imbalance with less sensitivity to hyperparameters \cite{chen2023class,qi2022novel,yan2025empowering,li2024cross}. Moreover, extending \methodname{} to more complex scenarios \cite{10540001,ijcai2024p478,yang2022learning,ijcai2025p177,wang2024causal} is a promising direction.

\section*{Acknowledgments}
This work is supported in part by the Key Research and Development Program of Shandong Province (Grant No. 2024TSGC0667) and the Ministry of Education, Singapore, under its Academic Research Fund Tier 1 (RG101/24).

{
\small
\bibliographystyle{unsrt}
\bibliography{NIPS2025references}

@article{krizhevsky2009learning,
  title={Learning multiple layers of features from tiny images},
  author={Krizhevsky, Alex and Hinton, Geoffrey and others},
  year={2009},
  publisher={Toronto, ON, Canada}
}

@article{le2015tiny,
  title={Tiny imagenet visual recognition challenge},
  author={Le, Ya and Yang, Xuan},
  journal={CS 231N},
  volume={7},
  number={7},
  pages={3},
  year={2015}
}

@article{kirkpatrick2017overcoming,
  title={Overcoming catastrophic forgetting in neural networks},
  author={Kirkpatrick, James and Pascanu, Razvan and Rabinowitz, Neil and Veness, Joel and Desjardins, Guillaume and Rusu, Andrei A and Milan, Kieran and Quan, John and Ramalho, Tiago and Grabska-Barwinska, Agnieszka and others},
  journal={Proceedings of the national academy of sciences},
  volume={114},
  number={13},
  pages={3521--3526},
  year={2017},
  publisher={National Academy of Sciences}
}

@article{li2017learning,
  title={Learning without forgetting},
  author={Li, Zhizhong and Hoiem, Derek},
  journal={IEEE transactions on pattern analysis and machine intelligence},
  volume={40},
  number={12},
  pages={2935--2947},
  year={2017},
  publisher={IEEE}
}

@inproceedings{zhang2023target,
  title={Target: Federated class-continual learning via exemplar-free distillation},
  author={Zhang, Jie and Chen, Chen and Zhuang, Weiming and Lyu, Lingjuan},
  booktitle={Proceedings of the IEEE/CVF International Conference on Computer Vision},
  pages={4782--4793},
  year={2023}
}

@inproceedings{tran2024text,
  title={Text-enhanced data-free approach for federated class-incremental learning},
  author={Tran, Minh-Tuan and Le, Trung and Le, Xuan-May and Harandi, Mehrtash and Phung, Dinh},
  booktitle={Proceedings of the IEEE/CVF Conference on Computer Vision and Pattern Recognition},
  pages={23870--23880},
  year={2024}
}

@inproceedings{li2024towards,
  title={Towards efficient replay in federated incremental learning},
  author={Li, Yichen and Li, Qunwei and Wang, Haozhao and Li, Ruixuan and Zhong, Wenliang and Zhang, Guannan},
  booktitle={Proceedings of the IEEE/CVF Conference on Computer Vision and Pattern Recognition},
  pages={12820--12829},
  year={2024}
}

@article{li2025re,
  title={Re-Fed+: A Better Replay Strategy for Federated Incremental Learning},
  author={Li, Yichen and Wang, Haozhao and Qi, Yining and Liu, Wei and Li, Ruixuan},
  journal={IEEE Transactions on Pattern Analysis and Machine Intelligence},
  year={2025},
  publisher={IEEE}
}

@inproceedings{chai2024efficient,
  title={Efficient decentralized federated singular vector decomposition},
  author={Chai, Di and Zhang, Junxue and Yang, Liu and Jin, Yilun and Wang, Leye and Chen, Kai and Yang, Qiang},
  booktitle={2024 USENIX Annual Technical Conference (USENIX ATC 24)},
  pages={1029--1047},
  year={2024}
}

@article{drineas2012fast,
  title={Fast approximation of matrix coherence and statistical leverage},
  author={Drineas, Petros and Magdon-Ismail, Malik and Mahoney, Michael W and Woodruff, David P},
  journal={The Journal of Machine Learning Research},
  volume={13},
  number={1},
  pages={3475--3506},
  year={2012},
  publisher={JMLR. org}
}

@inproceedings{hu2024fedmut,
  title={FedMut: Generalized Federated Learning via Stochastic Mutation},
  author={Hu, Ming and Cao, Yue and Li, Anran and Li, Zhiming and Liu, Chengwei and Li, Tianlin and Chen, Mingsong and Liu, Yang},
  booktitle={Proceedings of the AAAI Conference on Artificial Intelligence},
  volume={38},
  number={11},
  pages={12528--12537},
  year={2024}
}

@inproceedings{haozhao1,
author = {Haozhao Wang and Haoran Xu and Yichen Li and Yuan Xu and Ruixuan Li and Tianwei Zhang},
title = {FedCDA: Federated Learning with Cross-rounds Divergence-aware Aggregation},
booktitle = {The Twelfth International Conference on Learning Representations,
{ICLR} 2024, Vienna, Austria, May 7-11, 2024}
}

@inproceedings{qi2023cross,
  title={Cross-silo prototypical calibration for federated learning with non-iid data},
  author={Qi, Zhuang and Meng, Lei and Chen, Zitan and Hu, Han and Lin, Hui and Meng, Xiangxu},
  booktitle={Proceedings of the 31st ACM International Conference on Multimedia},
  pages={3099--3107},
  year={2023}
}

@article{fu2025learn,
  title={Learn the global prompt in the low-rank tensor space for heterogeneous federated learning},
  author={Fu, Lele and Huang, Sheng and Li, Yuecheng and Chen, Chuan and Zhang, Chuanfu and Zheng, Zibin},
  journal={Neural Networks},
  volume={187},
  pages={107319},
  year={2025},
  publisher={Elsevier}
}

@inproceedings{hu2024fedcross,
  title={FedCross: Towards Accurate Federated Learning via Multi-Model Cross-Aggregation},
  author={Hu, Ming and Zhou, Peiheng and Yue, Zhihao and Ling, Zhiwei and Huang, Yihao and Li, Anran and Liu, Yang and Lian, Xiang and Chen, Mingsong},
  booktitle={IEEE International Conference on Data Engineering (ICDE)},
  pages={2137--2150},
  year={2024},
  organization={IEEE}
}

@article{wu2025federated,
  title={Federated Class-Incremental Learning via Weighted Aggregation and Distillation},
  author={Wu, Feng and Tan, Alysa Ziying and Feng, Siwei and Yu, Han and Deng, Tao and Zhao, Libang and Chen, Yuanlu},
  journal={IEEE Internet of Things Journal},
  year={2025},
  publisher={IEEE}
}

@article{lu2024federated,
  title={Federated class-incremental learning with dynamic feature extractor fusion},
  author={Lu, Yanyan and Yang, Lei and Chen, Hao-Rui and Cao, Jiannong and Lin, Wanyu and Long, Saiqin},
  journal={IEEE Transactions on Mobile Computing},
  year={2024},
  publisher={IEEE}
}

@article{chen2024general,
  title={General federated class-incremental learning with lightweight generative replay},
  author={Chen, Yuanlu and Tan, Alysa Ziying and Feng, Siwei and Yu, Han and Deng, Tao and Zhao, Libang and Wu, Feng},
  journal={IEEE Internet of Things Journal},
  year={2024},
  publisher={IEEE}
}

@inproceedings{dong2022federated,
  title={Federated class-incremental learning},
  author={Dong, Jiahua and Wang, Lixu and Fang, Zhen and Sun, Gan and Xu, Shichao and Wang, Xiao and Zhu, Qi},
  booktitle={Proceedings of the IEEE/CVF conference on computer vision and pattern recognition},
  pages={10164--10173},
  year={2022}
}

@inproceedings{gao2024fedprok,
  title={Fedprok: Trustworthy federated class-incremental learning via prototypical feature knowledge transfer},
  author={Gao, Xin and Yang, Xin and Yu, Hao and Kang, Yan and Li, Tianrui},
  booktitle={Proceedings of the IEEE/CVF Conference on Computer Vision and Pattern Recognition},
  pages={4205--4214},
  year={2024}
}

@inproceedings{yoo2024federated,
  title={Federated Class Incremental Learning: A Pseudo Feature Based Approach Without Exemplars},
  author={Yoo, Min Kyoon and Park, Yu Rang},
  booktitle={Proceedings of the Asian Conference on Computer Vision},
  pages={488--498},
  year={2024}
}

@article{wang2024data,
  title={Data-Free Federated Class Incremental Learning with Diffusion-Based Generative Memory},
  author={Wang, Naibo and Deng, Yuchen and Feng, Wenjie and Yin, Jianwei and Ng, See-Kiong},
  journal={arXiv preprint arXiv:2405.17457},
  year={2024}
}

@inproceedings{liang2024diffusion,
  title={Diffusion-driven data replay: A novel approach to combat forgetting in federated class continual learning},
  author={Liang, Jinglin and Zhong, Jin and Gu, Hanlin and Lu, Zhongqi and Tang, Xingxing and Dai, Gang and Huang, Shuangping and Fan, Lixin and Yang, Qiang},
  booktitle={European Conference on Computer Vision},
  pages={303--319},
  year={2024},
  organization={Springer}
}

@article{babakniya2023data,
  title={A data-free approach to mitigate catastrophic forgetting in federated class incremental learning for vision tasks},
  author={Babakniya, Sara and Fabian, Zalan and He, Chaoyang and Soltanolkotabi, Mahdi and Avestimehr, Salman},
  journal={Advances in Neural Information Processing Systems},
  volume={36},
  pages={66408--66425},
  year={2023}
}

@inproceedings{rebuffi2017icarl,
  title={icarl: Incremental classifier and representation learning},
  author={Rebuffi, Sylvestre-Alvise and Kolesnikov, Alexander and Sperl, Georg and Lampert, Christoph H},
  booktitle={Proceedings of the IEEE conference on Computer Vision and Pattern Recognition},
  pages={2001--2010},
  year={2017}
}

@inproceedings{yu2024overcoming,
  title={Overcoming spatial-temporal catastrophic forgetting for federated class-incremental learning},
  author={Yu, Hao and Yang, Xin and Gao, Xin and Feng, Yihui and Wang, Hao and Kang, Yan and Li, Tianrui},
  booktitle={Proceedings of the 32nd ACM International Conference on Multimedia},
  pages={5280--5288},
  year={2024}
}

@inproceedings{tan2024fl,
  title={FL-Clip: Bridging Plasticity and Stability in Pre-Trained Federated Class-Incremental Learning Models},
  author={Tan, Alysa Ziying and Feng, Siwei and Yu, Han},
  booktitle={2024 IEEE International Conference on Multimedia and Expo (ICME)},
  pages={1--6},
  year={2024},
  organization={IEEE}
}

@inproceedings{psaltis2023fedrcil,
  title={Fedrcil: Federated knowledge distillation for representation based contrastive incremental learning},
  author={Psaltis, Athanasios and Chatzikonstantinou, Christos and Patrikakis, Charalampos Z and Daras, Petros},
  booktitle={Proceedings of the IEEE/CVF International Conference on Computer Vision},
  pages={3463--3472},
  year={2023}
}

@article{nguyen2024overcoming,
  title={Overcoming Catastrophic Forgetting in Federated Class-Incremental Learning via Federated Global Twin Generator},
  author={Nguyen, Thinh and Doan, Khoa D and Nguyen, Binh T and Le-Phuoc, Danh and Wong, Kok-Seng},
  journal={arXiv preprint arXiv:2407.11078},
  year={2024}
}

@inproceedings{qi2025cross,
  title={Cross-Silo Feature Space Alignment for Federated Learning on Clients with Imbalanced Data},
  author={Qi, Zhuang and Meng, Lei and et al.},
  booktitle={The 39th Annual AAAI Conference on Artificial Intelligence (AAAI-25)},
  pages={19986-19994},
  year={2025}
}

@article{qi2023better,
  title={Better generative replay for continual federated learning},
  author={Qi, Daiqing and Zhao, Handong and Li, Sheng},
  journal={arXiv preprint arXiv:2302.13001},
  year={2023}
}

@article{yang2024federated,
  title={Federated continual learning via knowledge fusion: A survey},
  author={Yang, Xin and Yu, Hao and Gao, Xin and Wang, Hao and Zhang, Junbo and Li, Tianrui},
  journal={IEEE Transactions on Knowledge and Data Engineering},
  volume={36},
  number={8},
  pages={3832--3850},
  year={2024},
  publisher={IEEE}
}

@article{chen2025knowledge,
  title={Knowledge Efficient Federated Continual Learning for Industrial Edge Systems},
  author={Chen, Jiao and He, Jiayi and Tang, Jianhua and Li, Weihua and Yin, Zihang},
  journal={IEEE Transactions on Network Science and Engineering},
  year={2025},
  publisher={IEEE}
}

@article{zhong2025sacfl,
  title={SacFL: Self-Adaptive Federated Continual Learning for Resource-Constrained End Devices},
  author={Zhong, Zhengyi and Bao, Weidong and Wang, Ji and Chen, Jianguo and Lyu, Lingjuan and Lim, Wei Yang Bryan},
  journal={IEEE Transactions on Neural Networks and Learning Systems},
  year={2025},
  publisher={IEEE}
}

@inproceedings{wang2024feddse,
  title={Feddse: Distribution-aware sub-model extraction for federated learning over resource-constrained devices},
  author={Wang, Haozhao and Jia, Yabo and Zhang, Meng and Hu, Qinghao and Ren, Hao and Sun, Peng and Wen, Yonggang and Zhang, Tianwei},
  booktitle={Proceedings of the ACM Web Conference 2024},
  pages={2902--2913},
  year={2024}
}

@inproceedings{wang2024fednlr,
  title={Fednlr: Federated learning with neuron-wise learning rates},
  author={Wang, Haozhao and Zheng, Peirong and Han, Xingshuo and Xu, Wenchao and Li, Ruixuan and Zhang, Tianwei},
  booktitle={Proceedings of the 30th ACM SIGKDD Conference on Knowledge Discovery and Data Mining},
  pages={3069--3080},
  year={2024}
}

@article{chen2024flexfl,
  title={Flexfl: Heterogeneous federated learning via apoz-guided flexible pruning in uncertain scenarios},
  author={Chen, Zekai and Jia, Chentao and Hu, Ming and Xie, Xiaofei and Li, Anran and Chen, Mingsong},
  journal={IEEE Transactions on Computer-Aided Design of Integrated Circuits and Systems},
  volume={43},
  number={11},
  pages={4069--4080},
  year={2024},
  publisher={IEEE}
}

@inproceedings{fu2025beyond,
  title={Beyond federated prototype learning: Learnable semantic anchors with hyperspherical contrast for domain-skewed data},
  author={Fu, Lele and Huang, Sheng and Lai, Yanyi and Liao, Tianchi and Zhang, Chuanfu and Chen, Chuan},
  booktitle={Proceedings of the AAAI Conference on Artificial Intelligence},
  volume={39},
  number={16},
  pages={16648--16656},
  year={2025}
}

@article{liao2025privacy,
  title={Privacy-preserving vertical federated learning with tensor decomposition for data missing features},
  author={Liao, Tianchi and Fu, Lele and Zhang, Lei and Yang, Lei and Chen, Chuan and Ng, Michael K and Huang, Huawei and Zheng, Zibin},
  journal={IEEE Transactions on Information Forensics and Security},
  year={2025},
  publisher={IEEE}
}

@inproceedings{fuless,
  title={Less is More: Federated Graph Learning with Alleviating Topology Heterogeneity from A Causal Perspective},
  author={Fu, Lele and Deng, Bowen and Huang, Sheng and Liao, Tianchi and Pan, Shirui and Chen, Chuan},
  booktitle={Forty-second International Conference on Machine Learning}
}

@ARTICLE{10323204,
  author={Dong, Jiahua and Li, Hongliu and Cong, Yang and Sun, Gan and Zhang, Yulun and Van Gool, Luc},
  journal={IEEE Transactions on Pattern Analysis and Machine Intelligence}, 
  title={No One Left Behind: Real-World Federated Class-Incremental Learning}, 
  year={2024},
  volume={46},
  number={4},
  pages={2054-2070},
  doi={10.1109/TPAMI.2023.3334213}}

@InProceedings{dong2023federated_FISS,
    author = {Dong, Jiahua and Zhang, Duzhen and Cong, Yang and Cong, Wei and Ding, Henghui and Dai, Dengxin},
    title = {Federated Incremental Semantic Segmentation},
    booktitle = {Proceedings of the IEEE/CVF Conference on Computer Vision and Pattern Recognition (CVPR)},
    month = {June},
    year = {2023},
    pages = {3934-3943}
}

@ARTICLE{10605121,
  author={Meng, Lei and Qi, Zhuang and Wu, Lei and Du, Xiaoyu and Li, Zhaochuan and Cui, Lizhen and Meng, Xiangxu},
  journal={IEEE Transactions on Neural Networks and Learning Systems}, 
  title={Improving Global Generalization and Local Personalization for Federated Learning}, 
  year={2025},
  volume={36},
  number={1},
  pages={76-87},
  doi={10.1109/TNNLS.2024.3417452}}

@article{li2024sr,
  title={Sr-fdil: Synergistic replay for federated domain-incremental learning},
  author={Li, Yichen and Xu, Wenchao and Wang, Haozhao and Qi, Yining and Li, Ruixuan and Guo, Song},
  journal={IEEE Transactions on Parallel and Distributed Systems},
  year={2024},
  publisher={IEEE}
}

@article{qi2025federated,
  title={Federated Deconfounding and Debiasing Learning for Out-of-Distribution Generalization},
  author={Qi, Zhuang and Zhou, Sijin and Meng, Lei and Hu, Han and Yu, Han and Meng, Xiangxu},
  journal={arXiv preprint arXiv:2505.04979},
  year={2025}
}

@article{qi2024cross,
  title={Cross-training with multi-view knowledge fusion for heterogenous federated learning},
  author={Qi, Zhuang and Meng, Lei and He, Weihao and Zhang, Ruohan and Wang, Yu and Qi, Xin and Meng, Xiangxu},
  journal={arXiv preprint arXiv:2405.20046},
  year={2024}
}

@inproceedings{feng2024federated,
  title={Federated multi-view clustering via tensor factorization},
  author={Feng, Wei and Wu, Zhenwei and Wang, Qianqian and Dong, Bo and Tao, Zhiqiang and Gao, Quanxue},
  booktitle={Proceedings of the Thirty-Third International Joint Conference on Artificial Intelligence, IJCAI-24},
  pages={3962--3970},
  year={2024}
}

@inproceedings{yu2011dynamic,
  title={Dynamic witness selection for trustworthy distributed cooperative sensing in cognitive radio networks},
  author={Yu, Han and Liu, Siyuan and Kot, Alex C and Miao, Chunyan and Leung, Cyril},
  booktitle={Proceedings of the 13th IEEE International Conference on Communication Technology (ICCT'11)},
  pages={1--6},
  year={2011}
}

@inproceedings{pan2016efficient,
  title={Efficient Collaborative Crowdsourcing},
  author={Pan, Zhengxiang and Yu, Han and Miao, Chunyan and Leung, Cyril},
  booktitle={The 30th AAAI Conference on Artificial Intelligence (AAAI-16)},
  pages={4248--4249},
  year={2016}
}

@inproceedings{an2022numerical,
  title={A numerical DEs perspective on unfolded linearized admm networks for inverse problems},
  author={An, Weixin and Yue, Yingjie and Liu, Yuanyuan and Shang, Fanhua and Liu, Hongying},
  booktitle={Proceedings of the 30th ACM International Conference on Multimedia},
  pages={5065--5073},
  year={2022}
}

@inproceedings{FedSSA,
  author    = {Liping Yi and Han Yu and Zhuan Shi and Gang Wang and Xiaoguang Liu and Lizhen Cui and Xiaoxiao Li},
  title     = {{FedSSA: Semantic Similarity-based Aggregation for Efficient Model-Heterogeneous Personalized Federated Learning}},
  booktitle = {Proc. {IJCAI}},
  year      = {2024},
}

@inproceedings{FedGH,
  author       = {Liping Yi and
                  Gang Wang and
                  Xiaoguang Liu and
                  Zhuan Shi and
                  Han Yu},
  title        = {FedGH: Heterogeneous Federated Learning with Generalized Global Header},
  booktitle    = {Proc.
                  {MM}, Ottawa, ON, Canada},
  pages        = {8686--8696},
  publisher    = {{ACM}},
  year         = {2023},
}

@inproceedings{wbk_mm2017,
  author       = {Bokun Wang and
                  Yang Yang and
                  Xing Xu and
                  Alan Hanjalic and
                  Heng Tao Shen},
  title        = {Adversarial Cross-Modal Retrieval},
  booktitle    = {Proceedings of the 2017 {ACM} on Multimedia Conference},
  pages        = {154--162},
  year         = {2017}
}

@article{xu2025self,
  title={Self-Reinforcing Prototype Evolution with Dual-Knowledge Cooperation for Semi-Supervised Lifelong Person Re-Identification},
  author={Xu, Kunlun and Zhuo, Fan and Li, Jiangmeng and Zou, Xu and Zhou, Jiahuan},
  journal={ICCV},
  year={2025},
  publisher={IEEE}
}

@article{xu2025long,
  title={Long Short-Term Knowledge Decomposition and Consolidation for Lifelong Person Re-Identification},
  author={Xu, Kunlun and Liu, Zichen and Zou, Xu and Peng, Yuxin and Zhou, Jiahuan},
  journal={TPAMI},
  year={2025},
  publisher={IEEE}
}

@inproceedings{xu2024distribution,
  title={Distribution-aware knowledge prototyping for non-exemplar lifelong person re-identification},
  author={Xu, Kunlun and Zou, Xu and Peng, Yuxin and Zhou, Jiahuan},
  booktitle={CVPR},
  pages={16604--16613},
  year={2024}
}

@inproceedings{liang2025tta,
  title={TTA-FedDG: Leveraging Test-Time Adaptation to Address Federated Domain Generalization},
  author={Liang, Haoyuan and Zhang, Xinyu and Cao, Shilei and Li, Guowen and Zheng, Juepeng},
  booktitle={Proceedings of the AAAI Conference on Artificial Intelligence},
  volume={39},
  number={18},
  pages={18658--18666},
  year={2025}
}

@inproceedings{chen2023class,
  title={Class-level structural relation modeling and smoothing for visual representation learning},
  author={Chen, Zitan and Qi, Zhuang and Cao, Xiao and Li, Xiangxian and Meng, Xiangxu and Meng, Lei},
  booktitle={Proceedings of the 31st ACM International Conference on Multimedia},
  pages={2964--2972},
  year={2023}
}

@article{qi2022novel,
  title={A novel density-based outlier detection method using key attributes},
  author={Qi, Zhuang and Chen, Xiaming},
  journal={Intelligent Data Analysis},
  volume={26},
  number={6},
  pages={1431--1449},
  year={2022},
  publisher={SAGE Publications Sage UK: London, England}
}

@article{yan2025empowering,
  title={Empowering Vision Transformers with Multi-Scale Causal Intervention for Long-Tailed Image Classification},
  author={Yan, Xiaoshuo and Li, Zhaochuan and Meng, Lei and Qi, Zhuang and Wu, Wei and Li, Zixuan and Meng, Xiangxu},
  journal={arXiv preprint arXiv:2505.08173},
  year={2025}
}

@article{li2024cross,
  title={Cross-modal learning using privileged information for long-tailed image classification},
  author={Li, Xiangxian and Zheng, Yuze and Ma, Haokai and Qi, Zhuang and Meng, Xiangxu and Meng, Lei},
  journal={Computational Visual Media},
  volume={10},
  number={5},
  pages={981--992},
  year={2024},
  publisher={TUP}
}

@inproceedings{qi2024attentive,
  title={Attentive modeling and distillation for out-of-distribution generalization of federated learning},
  author={Qi, Zhuang and He, Weihao and Meng, Xiangxu and Meng, Lei},
  booktitle={2024 IEEE International Conference on Multimedia and Expo (ICME)},
  pages={1--6},
  year={2024},
  organization={IEEE}
}

@inproceedings{ijcai2024p478,
  title     = {Exploiting Multi-Label Correlation in Label Distribution Learning},
  author    = {Kou, Zhiqiang and Wang, Jing and Tang, Jiawei and Jia, Yuheng and Shi, Boyu and Geng, Xin},
  booktitle = {Proceedings of the Thirty-Third International Joint Conference on
               Artificial Intelligence, {IJCAI-24}},
  publisher = {International Joint Conferences on Artificial Intelligence Organization},
  editor    = {Kate Larson},
  pages     = {4326--4334},
  year      = {2024},
  month     = {8},
  note      = {Main Track},
  doi       = {10.24963/ijcai.2024/478},
  url       = {https://doi.org/10.24963/ijcai.2024/478},
}

@ARTICLE{10540001,
  author={Kou, Zhiqiang and Wang, Jing and Jia, Yuheng and Geng, Xin},
  journal={IEEE Transactions on Circuits and Systems for Video Technology}, 
  title={Inaccurate Label Distribution Learning}, 
  year={2024},
  volume={34},
  number={10},
  pages={10237-10249},
  doi={10.1109/TCSVT.2024.3406066}}

@inproceedings{feng2025scalable,
  title={Scalable Federated One-Step Multi-View Clustering with Tensorized Regularization},
  author={Feng, Wei and Liu, Danting and Wang, Qianqian and Liang, Wenqi and Yan, Zheng},
  booktitle={Proceedings of the AAAI Conference on Artificial Intelligence},
  volume={39},
  number={16},
  pages={16586--16594},
  year={2025}
}

@inproceedings{yang2022learning,
  title={Learning with twin noisy labels for visible-infrared person re-identification},
  author={Yang, Mouxing and Huang, Zhenyu and Hu, Peng and Li, Taihao and Lv, Jiancheng and Peng, Xi},
  booktitle={Proceedings of the IEEE/CVF conference on computer vision and pattern recognition},
  pages={14308--14317},
  year={2022}
}

@inproceedings{ijcai2025p177,
  title     = {Learning Real Facial Concepts for Independent Deepfake Detection},
  author    = {Liu, Ming-Hui and Cheng, Harry and Wang, Tianyi and Luo, Xin and Xu, Xin-Shun},
  booktitle = {Proceedings of the International Joint Conference on Artificial Intelligence},
  pages     = {1585--1593},
  year      = {2025},
}

@inproceedings{zhong2025unlearning,
  title={Unlearning through knowledge overwriting: Reversible federated unlearning via selective sparse adapter},
  author={Zhong, Zhengyi and Bao, Weidong and Wang, Ji and Zhang, Shuai and Zhou, Jingxuan and Lyu, Lingjuan and Lim, Wei Yang Bryan},
  booktitle={Proceedings of the Computer Vision and Pattern Recognition Conference},
  pages={30661--30670},
  year={2025}
}

@article{zhong2022flee,
  title={Flee: A hierarchical federated learning framework for distributed deep neural network over cloud, edge, and end device},
  author={Zhong, Zhengyi and Bao, Weidong and Wang, Ji and Zhu, Xiaomin and Zhang, Xiongtao},
  journal={ACM Transactions on Intelligent Systems and Technology (TIST)},
  volume={13},
  number={5},
  pages={1--24},
  year={2022},
  publisher={ACM New York, NY}
}

@inproceedings{qi2022clustering,
  title={Clustering-based curriculum construction for sample-balanced federated learning},
  author={Qi, Zhuang and Wang, Yuqing and Chen, Zitan and Wang, Ran and Meng, Xiangxu and Meng, Lei},
  booktitle={CAAI international conference on artificial intelligence},
  pages={155--166},
  year={2022},
  organization={Springer}
}

@inproceedings{meng2025causal,
  title={Causal Inference over Visual-Semantic-Aligned Graph for Image Classification},
  author={Meng, Lei and Li, Xiangxian and Yan, Xiaoshuo and Ma, Haokai and Qi, Zhuang and Wu, Wei and Meng, Xiangxu},
  booktitle={Proceedings of the AAAI Conference on Artificial Intelligence},
  volume={39},
  number={18},
  pages={19449--19457},
  year={2025}
}

@article{wang2024causal,
  title={Causal inference for out-of-distribution recognition via sample balancing},
  author={Wang, Yuqing and Li, Xiangxian and Liu, Yannan and Cao, Xiao and Meng, Xiangxu and Meng, Lei},
  journal={CAAI Transactions on Intelligence Technology},
  volume={9},
  number={5},
  pages={1172--1184},
  year={2024},
  publisher={Wiley Online Library}
}
}




\newpage
\appendix

\section{Appendix}

\subsection{Experimental Results}
\subsubsection{Performance Comparison}
To thoroughly verify the effectiveness of the proposed \methodname{}, we compare its performance against various baselines under the setting of 10 clients. Based on the original implementations, we generate 10,240 synthetic samples per task for both TARGET and LANDER. The data replay configurations for Re-Fed and \methodname{} follow the settings outlined in Section~\ref{sec513}. The results are presented in Tables~\ref{tab6} and~\ref{tab7}. Consistent with the results shown in Tables~\ref{tab1} and~\ref{tab2}, \methodname{} achieves the best performance across all cases. Notably, \textbf{\methodname{} achieves over a 10\% gain compared to the second-best performing method in several settings.}

\begin{table*}[h]
\centering
\caption{Performance comparison between \methodname{} and baseline methods across CIFAR-10, CIFAR-100, and TinyImageNet under varying levels of data heterogeneity (Dirichlet parameter $\beta$). Specifically, CIFAR-10 is split into 3 tasks, CIFAR-100 into 5 tasks, and TinyImageNet into 10 tasks. The number of clients is fixed at 10, and all experiments are conducted with a random seed of 2023 to ensure reproducibility. The best results are \textbf{bolded}.}
\renewcommand{\arraystretch}{1.1}
\setlength{\tabcolsep}{8.pt}
\begin{tabular}{c|cc|ccc|ccc}
\toprule
\multirow{2}{*}{\textbf{Method}} & \multicolumn{2}{c|}{\textbf{CIFAR10}} & \multicolumn{3}{c|}{\textbf{CIFAR100}} & \multicolumn{3}{c}{\textbf{TinyImageNet}} \\\cline{2-9}
& $\beta{=}0.5$ & $\beta{=}1.0$ & $\beta{=}0.1$ & $\beta{=}0.5$ & $\beta{=}1.0$ & $\beta{=}0.1$ & $\beta{=}0.5$ & $\beta{=}1.0$  \\
\midrule
FedEWC & 36.40 & 42.00 & 15.19 & 18.66 & 19.50 & 6.19 & 7.23 & 7.78  \\
FedLwF & 48.24 & 49.11 & 27.02 & 37.92 & 41.77 & 10.67 & 13.02 & 14.73  \\
TARGET & 38.23 & 41.11 & 18.34 & 23.59 & 25.71 & 7.45 & 8.29 & 8.87  \\
LANDER & 41.54 & 45.52 & 30.83 & 43.69 & 47.29 & 12.33 & 15.18 & 15.64  \\
Re-Fed & 45.49 & 52.22 & 31.81 & 36.40 & 37.95 & 9.28 & 11.48 & 12.10 \\\hline
\textbf{\methodname{}} & 59.80 & 62.59 & 42.25 & 47.90 & 48.55 & 14.81 & 16.54 & 17.43  \\
\bottomrule
\end{tabular}
\label{tab6}
\end{table*}

\begin{table*}[h]
\centering
\caption{Performance comparison between \methodname{} and baseline methods across CIFAR-10, CIFAR-100, and TinyImageNet under varying levels of data heterogeneity (Dirichlet parameter $\beta$). Specifically, CIFAR-10 is split into 5 tasks, CIFAR-100 into 10 tasks, and TinyImageNet into 20 tasks. The number of clients is fixed at 10, and all experiments are conducted with a random seed of 2023 to ensure reproducibility. The best results are \textbf{bolded}.}
\renewcommand{\arraystretch}{1.1}
\setlength{\tabcolsep}{8.pt}
\begin{tabular}{c|cc|ccc|ccc}
\toprule
\multirow{2}{*}{\textbf{Method}} & \multicolumn{2}{c|}{\textbf{CIFAR10}} & \multicolumn{3}{c|}{\textbf{CIFAR100}} & \multicolumn{3}{c}{\textbf{TinyImageNet}} \\\cline{2-9}
& $\beta{=}0.5$ & $\beta{=}1.0$ & $\beta{=}0.1$ & $\beta{=}0.5$ & $\beta{=}1$ & $\beta{=}0.1$ & $\beta{=}0.5$ & $\beta{=}1.0$  \\
\midrule
FedEWC & 20.18 & 23.33 & 6.68 & 10.98 & 12.30 & 3.27 & 4.80 & 4.89 \\
FedLwF  & 43.31 & 46.79 & 13.82 & 17.79 & 27.80 & 4.50 & 5.71 & 9.07 \\
TARGET & 21.60 & 28.39 & 12.11 & 16.64 & 17.14 & 3.45 & 4.88 & 5.01 \\
LANDER & 27.24 & 32.21 & 10.74 & 25.87 & 31.79 & 4.74 & 12.05 & 13.21  \\
Re-Fed & 38.28 & 39.22 & 28.08 & 33.52 & 37.27 & 7.95 & 8.53 & 10.13 \\\hline
\textbf{\methodname{}} & 51.71 & 59.57 & 37.42 & 43.82 & 45.50 & 11.51 & 14.45 & 15.25 \\
\bottomrule
\end{tabular}
\label{tab7}
\end{table*}


\subsubsection{Performance Evaluation of \methodname{} under Incremental Tasks}
\begin{figure}[h]
\centering
\includegraphics[width=1.0\linewidth]{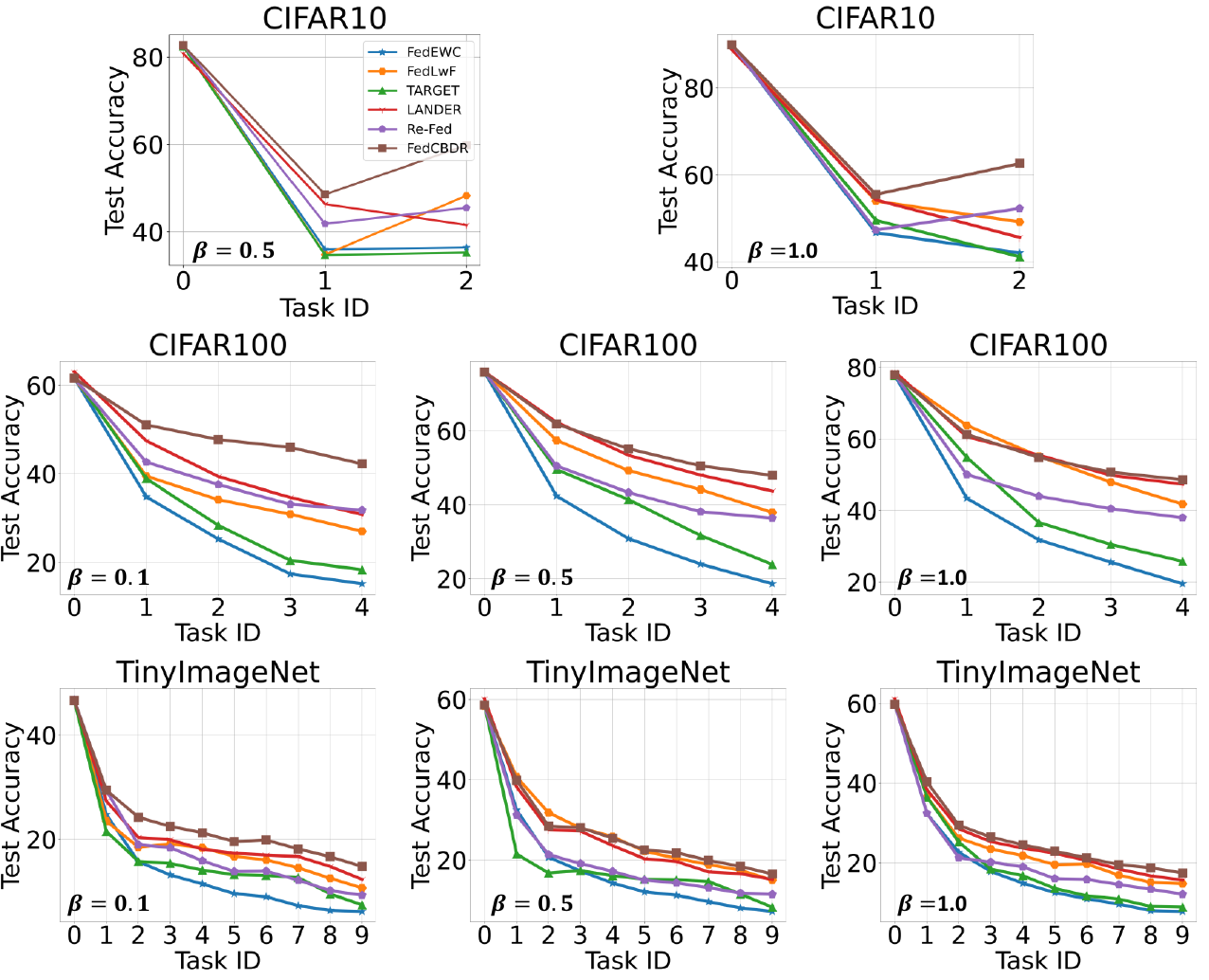}
\caption{Comparison of per-class sample distributions in the replay buffer between \methodname{} and Re-Fed on the CIFAR100 dataset, conducted under a heterogeneity level of $\beta = 0.5$, with a 10-task split and 5 clients.
} 
\label{fig7}
\end{figure}

We evaluate the performance evolution of \methodname{} and competing methods under a 10-client setting across incremental tasks on three benchmark datasets. Specifically, CIFAR-10 is split into 3 tasks ($\beta=\{0.5,1.0\}$), CIFAR-100 into 5 tasks ($\beta=\{0.1,0.5,1.0\}$), and TinyImageNet into 10 tasks ($\beta=\{0.1,0.5,1.0\}$). As shown in Figure \ref{fig7}, \textbf{\methodname{} consistently outperforms all baseline methods across incremental tasks, maintaining higher accuracy on both current and previous tasks throughout the training process}. Moreover, its performance degrades more slowly as the number of tasks increases.

\subsubsection{Comparison of Per-Class Sample Distributions in the Replay Buffer}
\begin{figure}[h]
\centering
\includegraphics[width=1.0\linewidth]{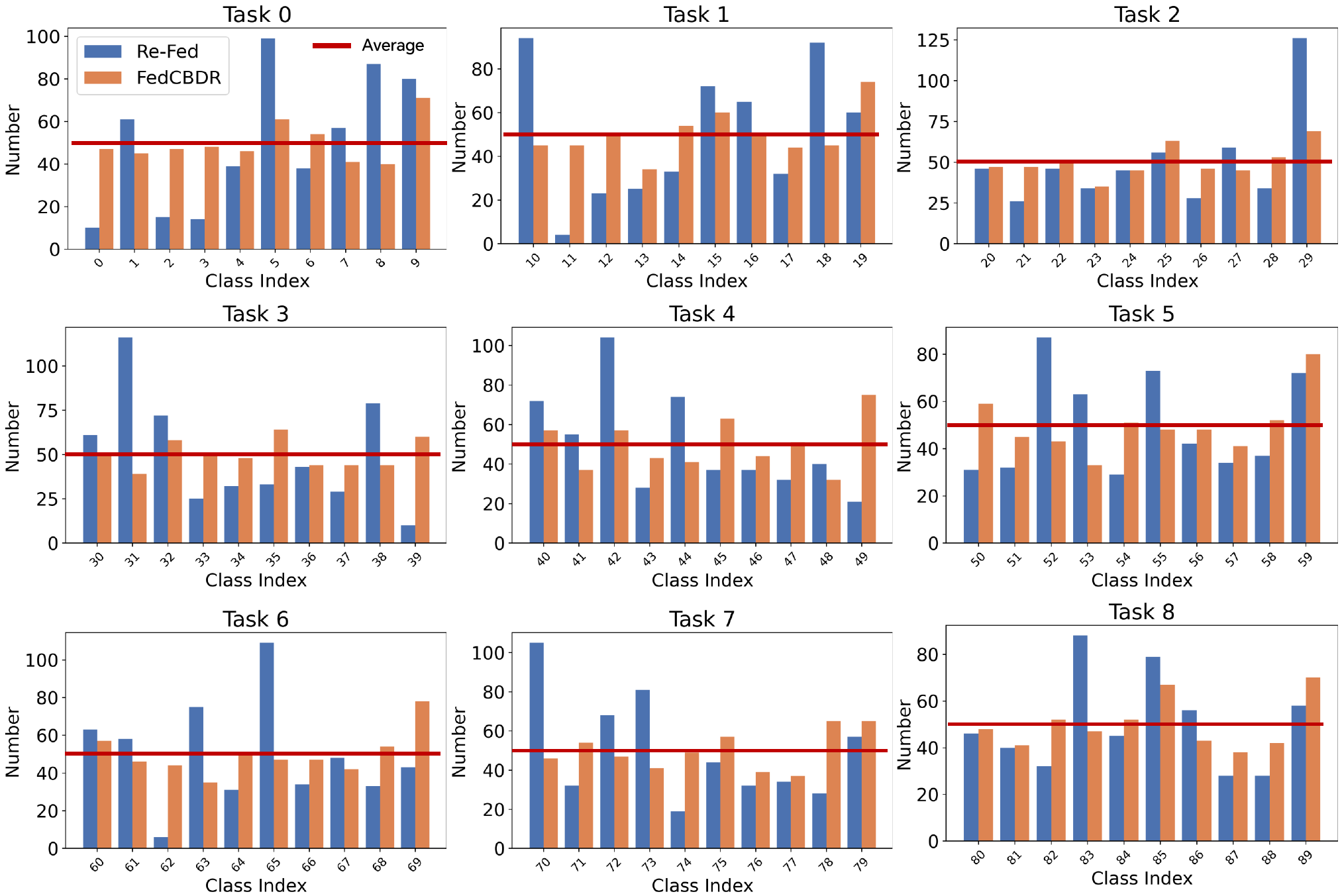}
\caption{Comparison of per-class sample distributions in the replay buffer between \methodname{} and Re-Fed on the CIFAR100 dataset, conducted under a heterogeneity level of $\beta = 0.5$, with a 10-task split and 5 clients.
} 
\label{fig8}
\end{figure}
We further validate the capability of the proposed \methodname{} to balance per-class sample distributions in more complex scenarios. Specifically, we divide the CIFAR100 dataset into 10 tasks. As illustrated in Figure \ref{fig8}, Re-Fed exhibits substantial disparities in the number of replayed samples across classes. For example, in task 1, while classes 10 and 18 contain nearly 100 samples each, class 11 has fewer than 10. In contrast, \textbf{\methodname{} effectively alleviates such class imbalance, with the number of replayed samples for all classes remaining consistently close to the average (as marked by the red line)}. This contributes to more stable knowledge retention across tasks and enhances overall model generalization.

\subsubsection{Quantitative Analysis of Replay Buffer Size on Test Accuracy}
\begin{table}[h]
\centering
\vspace{-0.5cm}
\caption{Comparison of model performance with varying replay budget $M$ per task across datasets, with the number of clients fixed at 10, and heterogeneity level $\beta=0.5$.}
\renewcommand{\arraystretch}{1}
\setlength{\tabcolsep}{8.2pt}
\begin{tabular}{c|ccc|ccc}
\toprule
\multirow{2}{*}{\textbf{Methods}} & \multicolumn{3}{c|}{\textbf{CIFAR10}} & \multicolumn{3}{c}{\textbf{CIFAR100}} \\
\cline{2-7}
 & M=150 & M=300 & M=450 & M=500 & M=1000 & M=1500 \\
\midrule
Re-Fed          & 39.22 & 42.93 & 45.49 & 28.21 & 36.40 & 41.78 \\
\methodname{}         & 48.15 & 54.62 & 59.80 & 38.34 & 47.90 & 51.14 \\
\bottomrule
\end{tabular}
\label{tab8}
\end{table}
We compare the performance of the data replay-based methods, Re-Fed and \methodname{}, under varying replay buffer budgets. Specifically, for CIFAR10, the buffer size is adjusted among $\{150, 300, 450\}$, while for CIFAR100, it ranges from $\{500, 1000, 1500\}$. The number of clients is set to 10, and heterogeneity level $\beta=0.5$. As shown in Table 8, \textbf{the performance of both methods improves as the buffer size increases, with \methodname{} maintaining a clear advantage over Re-Fed under all settings.} This also underscores the importance of balancing per-class sample counts in the replay buffer to ensure fair representation and stable performance.

\subsubsection{Evaluation on the Impact of Local Training Epochs}
\begin{figure}[h]
\centering
\includegraphics[width=1.0\linewidth]{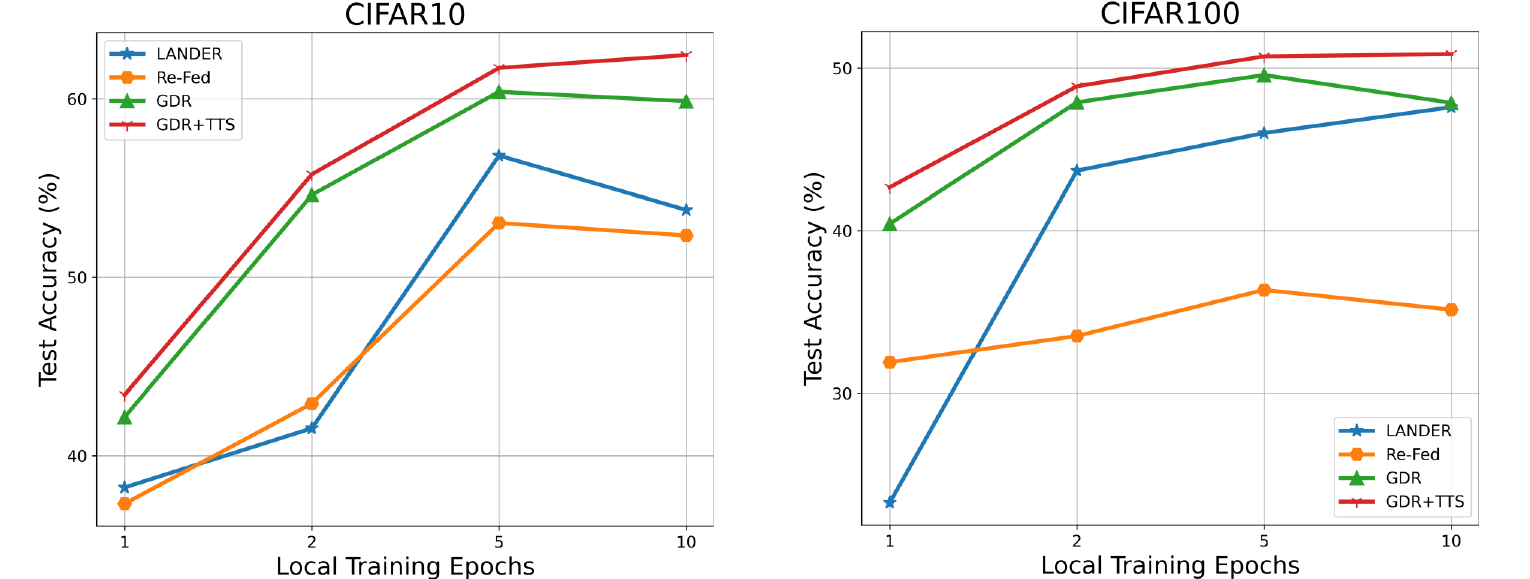}
\caption{Comparison of the final performance of the LANDER, Re-Fed, GDR, and GDR+TTS methods under different numbers of local training epochs. GDR and TTS are the two functional modules proposed in this work, and GDR+TTS=\methodname{}.
} 
\label{fig9}
\end{figure}

To assess the impact of local training intensity, we compare the performance of LANDER, Re-Fed, and \methodname{}, under varying local training epoch settings. Specifically, the evaluation is conducted on CIFAR10 divided into 3 tasks and CIFAR100 divided into 5 tasks, under a federated setting with 10 clients and a heterogeneity level of $\beta = 0.5$. As shown in Figure \ref{fig9}, \textbf{both GDR and GDR+TTS consistently outperform the baseline methods (LANDER and Re-Fed) across all local training epoch settings on both CIFAR10 and CIFAR100}. Moreover, \textbf{GDR+TTS achieves the highest test accuracy in every configuration}. The  improvement brought by TTS highlights its necessity in alleviating class imbalance during local training. And, unlike other methods whose performance drops at 10 local epochs due to biased updates, \textbf{GDR+TTS demonstrates a sustained improvement potential.}

\subsubsection{Performance Assessment of the Final Model Across Tasks}
This section compares the final model performance of different methods (LANDER, Re-Fed, \methodname{}) across various tasks. Specifically, all experiments are conducted under a federated setting with 5 clients and a heterogeneity level of $\beta = 0.5$. CIFAR10 is split into 3 tasks and CIFAR100 into 5 tasks. Each task is trained for 50 communication rounds, with each client performing 2 local training epochs per round using a batch size of 128. For sample replay, LANDER synthesizes 10,240 samples per task, while Re-Fed and \methodname{} retain 150 and 1,000 real samples per task on CIFAR10 and CIFAR100, respectively. As shown in Table \ref{tab9}, \textbf{LANDER suffers from significant forgetting of earlier tasks}, as evidenced by its low accuracy of only 1.37\% on Task 1 of CIFAR10. This indicates a severe inability to retain prior knowledge. Moreover, \textbf{LANDER also shows a noticeable decline in performance on the last task}, achieving only 57.00\% on Task 5 of CIFAR100, suggesting that its generalization to new tasks is also limited under non-i.i.d. conditions. Compared to LANDER and Re-Fed, \textbf{GDR significantly enhances the retention of knowledge from most early tasks}. \textbf{This demonstrates the advantage of balanced sample replay over imbalanced sampling}. In particular, \textbf{GDR+TTS outperforms GDR alone, highlighting the effectiveness of the proposed TTS module in mitigating class imbalance} and supporting long-term knowledge preservation under non-i.i.d. settings.

\begin{table}[ht]
\centering
\caption{Per-task and average accuracy (\%) of different methods on CIFAR10 and CIFAR100.}
\renewcommand{\arraystretch}{1.3}
\setlength{\tabcolsep}{3.5pt}
\begin{tabular}{l|ccc|c|ccccc|c}
\hline
\multirow{2}{*}{} & \multicolumn{4}{c|}{CIFAR10} & \multicolumn{6}{c}{CIFAR100} \\
\cline{2-11}
& Task 1 & Task 2 & Task 3 & ALL & Task 1 & Task 2 & Task 3 & Task 4 & Task 5 & ALL \\
\hline
LANDER & 1.37 & 30.00 & 88.32 & 44.74 & 33.95 & 40.90 & 43.70 & 44.45 & 57.00 & 44.00 \\
Re-Fed & 14.20 & 18.33 & 95.88 & 44.28 & 23.40 & 22.00 & 21.10 & 34.10 & 81.70 & 36.46 \\
GDR & 17.07 & 19.00 & 96.10 & 49.26 & 40.40 & 37.90 & 39.25 & 47.50 & 80.45 & 49.10 \\
GDR+TTS & 21.43 & 21.46 & 96.08 & 51.30 & 41.45 & 38.05 & 38.50 & 48.55 & 81.40 & 49.59 \\
\hline
\end{tabular}
\label{tab9}
\end{table}

\subsubsection{Scalability, Complexity, and Communication Efficiency}
 We have conducted additional large-scale experiments by extending the number of clients to 50/100 while simulating asynchronous participation, and the client sampling rate is set to 0.2. The dataset is divided into five tasks, and in Re-Fed and FedCBDR, the number of replay samples is set to 150/300 for CIFAR-10 and 500/1000 for CIFAR-100.The following results can be summarized:
\begin{table}[h]
\centering
\caption{Comparison of continual federated learning methods with 50 clients on CIFAR10 and CIFAR100 under different Dirichlet data partitions ($\alpha$).}
\setlength{\tabcolsep}{6pt}
\renewcommand{\arraystretch}{1.15}
\begin{tabular}{lcccc}
\toprule
\multirow{2}{*}{\textbf{50 Clients}} &
\multicolumn{2}{c}{CIFAR10} &
\multicolumn{2}{c}{CIFAR100} \\
\cmidrule(lr){2-3}\cmidrule(lr){4-5}
& $\alpha=0.5$ & $\alpha=1.0$ & $\alpha=0.5$ & $\alpha=1.0$ \\
\midrule
FedLwF              & 17.51 & 19.43 & 19.99 & 23.91 \\
LANDER              & 18.08 & 21.27 & 26.96 & 27.34 \\
Re-Fed (300/500)    & 29.65 & 32.45 & 27.12 & 28.86 \\
FedCBDR (300/500)   & 34.76 & 35.98 & 28.75 & 30.90 \\
Re-Fed (600/1000)   & 36.40 & 38.46 & 35.94 & 37.79 \\
FedCBDR (600/1000)  & \textbf{41.33} & \textbf{45.31} & \textbf{38.54} & \textbf{39.69} \\
\bottomrule
\end{tabular}
\label{tab10}
\end{table}

Moreover, we have further clarified the computational complexity of the SVD step.
Client-side ISVD costs $\mathcal{O}(n_k d^{2})$ per client, while server-side SVD costs
$\mathcal{O}(N d^{2})$, where $N=\sum_{k} n_k$ is the total number of pseudo-features.

\medskip
\noindent
We also provide a summary of the additional training cost on clients, the communication overhead,
and the server-side computation incurred by our method in comparison to FedAvg.
\textbf{Overall, the additional cost of performing SVD and computing leverage scores is relatively small.}

\begin{table}[t]
\centering
\caption{Computation and communication analysis on CIFAR10 and CIFAR100 datasets.
Top: computation overhead per client; Bottom: communication cost per round.}
\setlength{\tabcolsep}{2pt}
\renewcommand{\arraystretch}{1.15}

\begin{tabular}{
    p{1.8cm}  
    p{1.2cm}  
    p{1.7cm}  
    p{1.7cm}  
    p{2.0cm}  
    p{2.5cm}  
    p{2.5cm}  
}
\toprule
\textbf{Dataset} & \textbf{\#Clients (K)} & \textbf{Samples/ Client ($n_k$)} & \textbf{Feature Dim ($d$)} &
\textbf{ISVD Time/Client (s)} & \textbf{Feature Extraction Time/Client (s)} & \textbf{Total Extra Time/Client (s)} \\
\midrule
CIFAR10  & 5 & 2000 & 512 & 0.00580 & 0.9083 & 0.9141 \\
CIFAR100 & 5 & 2000 & 512 & 0.00452 & 1.0250 & 1.2952 \\
\bottomrule
\end{tabular}

\vspace{1em}

\begin{tabular}{
    p{1.8cm}
    p{1.2cm}
    p{1.7cm}
    p{1.7cm}
    p{2.5cm}
    p{2.7cm}
    p{2.3cm}
}
\toprule
\textbf{Dataset} & \textbf{\#Clients (K)} & \textbf{Samples/ Client ($n_k$)} & \textbf{Feature Dim ($d$)} &
\textbf{Upload/Client (MB)} & \textbf{Total Upload per Round (MB)} & \textbf{Index Download (KB)} \\
\midrule
CIFAR10 & 10 & 100 & 512 & 3.906  & 19.53  & 39.06 \\
CIFAR10 & 10 & 100 & 128 & 0.9765 & 4.8825 & 39.06 \\
CIFAR10 & 10 & 100 & 32  & 0.2441 & 1.2206 & 39.06 \\
\bottomrule
\end{tabular}
\label{tab11}
\end{table}


\newpage
\section*{NeurIPS Paper Checklist}

\begin{enumerate}

\item {\bf Claims}
    \item[] Question: Do the main claims made in the abstract and introduction accurately reflect the paper's contributions and scope?
    \item[] Answer: \answerYes{} 
    \item[] Justification: The research problem and the main contributions of this study are clearly articulated in the abstract and introduction sections.

\item {\bf Limitations}
    \item[] Question: Does the paper discuss the limitations of the work performed by the authors?
    \item[] Answer: \answerYes{} 
    \item[] Justification: The conclusion and future work section include a discussion of the study’s limitations.

\item {\bf Theory assumptions and proofs}
    \item[] Question: For each theoretical result, does the paper provide the full set of assumptions and a complete (and correct) proof?
    \item[] Answer: \answerNA{} 
    \item[] Justification: This paper does not involve theoretical assumptions

    \item {\bf Experimental result reproducibility}
    \item[] Question: Does the paper fully disclose all the information needed to reproduce the main experimental results of the paper to the extent that it affects the main claims and/or conclusions of the paper (regardless of whether the code and data are provided or not)?
    \item[] Answer: \answerYes{} 
    \item[] Justification: This paper provides a detailed description of the experimental setup, including the parameter tuning ranges, and the code will be made available as supplementary material.

\item {\bf Open access to data and code}
    \item[] Question: Does the paper provide open access to the data and code, with sufficient instructions to faithfully reproduce the main experimental results, as described in supplemental material?
    \item[] Answer: \answerYes{} 
    \item[] Justification: The code will be made available as supplementary material.

\item {\bf Experimental setting/details}
    \item[] Question: Does the paper specify all the training and test details (e.g., data splits, hyperparameters, how they were chosen, type of optimizer, etc.) necessary to understand the results?
    \item[] Answer: \answerYes{} 
    \item[] Justification: The experimental setup is clearly detailed in both the main experimental section and the appendix.

\item {\bf Experiment statistical significance}
    \item[] Question: Does the paper report error bars suitably and correctly defined or other appropriate information about the statistical significance of the experiments?
    \item[] Answer: \answerYes{} 
    \item[] Justification: This paper reports the mean and standard deviation over multiple runs.

\item {\bf Experiments compute resources}
    \item[] Question: For each experiment, does the paper provide sufficient information on the computer resources (type of compute workers, memory, time of execution) needed to reproduce the experiments?
    \item[] Answer: \answerYes{} 
    \item[] Justification: The computational resources utilized are described in the experimental implementation details section.
    
\item {\bf Code of ethics}
    \item[] Question: Does the research conducted in the paper conform, in every respect, with the NeurIPS Code of Ethics \url{https://neurips.cc/public/EthicsGuidelines}?
    \item[] Answer: \answerYes{} 
    \item[] Justification: This work fully adheres to the NeurIPS Code of Ethics in all aspects.

\item {\bf Broader impacts}
    \item[] Question: Does the paper discuss both potential positive societal impacts and negative societal impacts of the work performed?
    \item[] Answer: \answerYes{} 
    \item[] Justification: The introduction highlights the significance of multi-source collaborative modeling.
    
\item {\bf Safeguards}
    \item[] Question: Does the paper describe safeguards that have been put in place for responsible release of data or models that have a high risk for misuse (e.g., pretrained language models, image generators, or scraped datasets)?
    \item[] Answer: \answerNA{} 
    \item[] Justification: The ResNet model and datasets used in this study are all open-source.

\item {\bf Licenses for existing assets}
    \item[] Question: Are the creators or original owners of assets (e.g., code, data, models), used in the paper, properly credited and are the license and terms of use explicitly mentioned and properly respected?
    \item[] Answer: \answerYes{} 
    \item[] Justification: All relevant works are properly cited, and all open-source assets are used in accordance with their licensing terms.

\item {\bf New assets}
    \item[] Question: Are new assets introduced in the paper well documented and is the documentation provided alongside the assets?
    \item[] Answer: \answerYes{} 
    \item[] Justification: An anonymized version of the code developed in this study is included in the supplementary materials to ensure reproducibility.

\item {\bf Crowdsourcing and research with human subjects}
    \item[] Question: For crowdsourcing experiments and research with human subjects, does the paper include the full text of instructions given to participants and screenshots, if applicable, as well as details about compensation (if any)? 
    \item[] Answer: \answerNA{} 
    \item[] Justification: The paper does not involve crowdsourcing nor research with human subjects.

\item {\bf Institutional review board (IRB) approvals or equivalent for research with human subjects}
    \item[] Question: Does the paper describe potential risks incurred by study participants, whether such risks were disclosed to the subjects, and whether Institutional Review Board (IRB) approvals (or an equivalent approval/review based on the requirements of your country or institution) were obtained?
    \item[] Answer: \answerNA{} 
    \item[] Justification: The paper does not involve crowdsourcing nor research with human subjects.

\item {\bf Declaration of LLM usage}
    \item[] Question: Does the paper describe the usage of LLMs if it is an important, original, or non-standard component of the core methods in this research? Note that if the LLM is used only for writing, editing, or formatting purposes and does not impact the core methodology, scientific rigorousness, or originality of the research, declaration is not required.
    \item[] Answer: \answerNA{} 
    \item[] Justification: LLMs were used only for grammar correction and  refinement.

\end{enumerate}

\end{document}